\documentclass[lettersize,journal]{IEEEtran}
\usepackage{amsmath,amsfonts}

\usepackage{multirow}
\usepackage{booktabs}
\usepackage[misc]{ifsym}
\usepackage{ntheorem}
\usepackage{float}
\usepackage{subfigure}
\usepackage{threeparttable}

\usepackage{algorithmic}
\usepackage{algorithm}
\usepackage{array}
\usepackage[caption=false,font=normalsize,labelfont=sf,textfont=sf]{subfig}
\usepackage{textcomp}
\usepackage{stfloats}
\usepackage{url}
\usepackage{verbatim}
\usepackage{graphicx}
\usepackage{cite}
\usepackage{caption}
\usepackage{colortbl}
\usepackage{diagbox}
\definecolor{mygray}{gray}{.7}

\def\eg{{\em e.g.}}
\def\etal{{\em et al.}}

\graphicspath{{./fig/}}

\newcommand{\myPara}[1]{\vspace{.05in}\noindent\textbf{#1}}

\newcommand{\mc}[1]{\mathcal{#1}}

\newcommand{\bm}[1]{\mbox{\boldmath{$#1$}}}

{}
{}
{}

\usepackage{graphicx}
\usepackage{float}
\usepackage{subfig}

\usepackage{xcolor}
\usepackage{xpatch}

\makeatother

\begin{document}

\title{Low-Resolution Object Recognition with Cross-Resolution Relational Contrastive Distillation}

\author{Kangkai Zhang,
        Shiming~Ge,~\IEEEmembership{Senior Member,~IEEE,}
        Ruixin Shi,
        and~Dan~Zeng,~\IEEEmembership{Senior Member,~IEEE}
\thanks{Kangkai Zhang is with the Institute of Information Engineering, Chinese Academy of Sciences, Beijing 100084, China, and with Baidu Inc., Beijing 100080, China. Email: zhangkangkai99@gmail.com.}
\thanks{Shiming Ge and Ruixin Shi are with the Institute of Information Engineering, Chinese Academy of Sciences, Beijing 100084, China, and with School of Cyber Security at University of Chinese Academy of Sciences, Beijing 100049, China. Email: \{geshiming, shiruixin\}@iie.ac.cn.}
\thanks{Dan~Zeng is with the Department of Communication Engineering, Shanghai
University, Shanghai 200040, China. E-mail: dzeng@shu.edu.cn.}
\thanks{Shiming Ge is the responding author. Email: geshiming@iie.ac.cn.}}

\markboth{IEEE Transactions on Circuits and Systems for Video Technology}%
{Shell \MakeLowercase{\textit{et al.}}: A Sample Article Using IEEEtran.cls for IEEE Journals}


\maketitle

\begin{abstract}
Recognizing objects in low-resolution images is a challenging task due to the lack of informative details. Recent studies have shown that knowledge distillation approaches can effectively transfer knowledge from a high-resolution teacher model to a low-resolution student model by aligning cross-resolution representations. However, these approaches still face limitations in adapting to the situation where the recognized objects exhibit significant representation discrepancies between training and testing images. In this study, we propose a cross-resolution relational contrastive distillation approach to facilitate low-resolution object recognition. Our approach enables the student model to mimic the behavior of a well-trained teacher model which delivers high accuracy in identifying high-resolution objects. To extract sufficient knowledge, the student learning is supervised with contrastive relational distillation loss, which preserves the similarities in various relational structures in contrastive representation space. In this manner, the capability of recovering missing details of familiar low-resolution objects can be effectively enhanced, leading to a better knowledge transfer. Extensive experiments on low-resolution object classification and low-resolution face recognition clearly demonstrate the effectiveness and adaptability of our approach.
\end{abstract}

\begin{IEEEkeywords}
Low-resolution face recognition, low-resolution object classification, knowledge distillation, domain adaptation.
\end{IEEEkeywords}

\section{Introduction}
\IEEEPARstart{W}{ith} the rapid development of deep learning, deep models have demonstrated remarkable success in various visual recognition applications~\cite{jia2021scaling,huang2022tcsvt,imagenet_cvpr09,ben2022tackling}. For example, EfficientNet~\cite{jia2021scaling} delivers a top-1 classification accuracy of 88.61\% on ImageNet~\cite{imagenet_cvpr09} in large-scale visual recognition, Groupface~\cite{Groupface} gives an extreme high accuracy of 99.85\% on LFW~\cite{LFWTech} in face verification, cross-domain methods deliver impressive performance in gait recognition ~\cite{huang2022tcsvt} and micro-expression recognition~\cite{ben2022tackling}. These achievements can be attributed to the ability of deep models with massive parameters to extract rich knowledge from extensive high-quality datasets. However, it may suffer from a sharp drop in accuracy when directly applying these models in practical scenarios due to domain distribution difference, i.e., the identified objects lack informative details due to occlusion~\cite{ge2020tcsvt} or low resolution~\cite{yan2022tcsvt}. Meanwhile, it is difficult to correct sufficient low-resolution training data in practical scenarios. Thus, it is necessary to explore a feasible solution that can address a key challenge in low-resolution object recognition: How to effectively transfer knowledge from high-resolution source domain to low-resolution target domain with minimal accuracy loss?

\begin{figure}[t]
\centering
\includegraphics[width=1.0\linewidth]{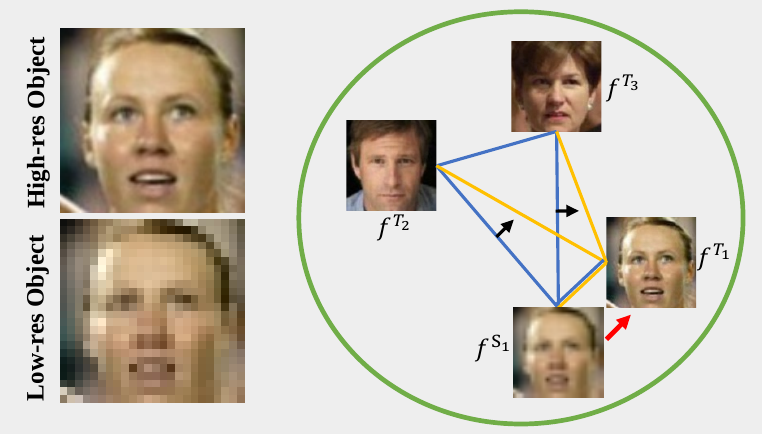}
\caption{A human who is more familiar with a high-resolution object can recognize the corresponding low-resolution one better. Transferring the structural relation knowledge between different resolution samples can help recognizing low-resolution objects. Our cross-resolution relational contrastive distillation enables low-resolution samples ($f^{S_i}$) to mimic the structural relation between corresponding high-resolution sample ($f^{T_i}$) and other high-resolution samples ($f^{T_j}, i\neq j$).
}
\label{fig:motivation}
\end{figure}

As shown in Fig.~\ref{fig:motivation}, in spite of the missing of many informative details, low-resolution objects still can be well recognized by subjects when they are familiar with the corresponding high-resolution objects. Recent works~\cite{ge2018low,GeHORKD20,ge2020efficient,beyer2022cvpr} have shown that it is feasible to improve the recognition capacity of a model by knowledge transfer from high-resolution domain to low-resolution one. According to the level of this cross-resolution knowledge transfer, current approaches can be mainly grouped into sample-level or relation-level approaches.
\IEEEpubidadjcol
For sample-level knowledge transfer, Wang \etal~\cite{wang2016studying} first proposed to use the corresponding high-resolution images to facilitate the model to extract features from low-resolution images. Subsequently, by learning low-resolution face representations and mimicking the adapted high-resolution knowledge, a light-weight student model can be constructed with high efficiency and promising accuracy in recognizi ng low-resolution faces~\cite{ge2018low,ge2020efficient}. However, sample-level knowledge is limited and insufficient to help the model extract sufficiently discriminative features, especially for cross-resolution knowledge transfer. Therefore, the researchers explored the relation-level knowledge transfer. Some recent works have shown that transferring the structural similarity instead of representation is beneficial to student learning~\cite{peng2019correlation,tung2019similarity,park2019relational,dong2021few}. Ge \etal~\cite{GeHORKD20} proposed a hybrid order relational distillation to distill richer knowledge from pretrained high-resolution models to facilitate low-resolution object recognition. In general, these approaches have achieved impressive performance. However, they all use low-order relation knowledge to model the mutual information, which may ignore complex high-order inter-sample interdependencies, \eg, contrastive relation, and lead to insufficient knowledge transfer for object recognition.

Recently, contrastive learning approaches \cite{oord2018representation,tian2019crd,xu2020knowledge,zhu2021complementary} have been widely used to learn feature representations from data samples by comparing the data with the positive and negative samples in the feature space. These approaches only need to learn discrimination in the feature space. Thus, they will not pay too much attention to pixel details, but can focus on more abstract semantic information, leading to simpler processing than pixel-level reconstruction~\cite{oord2018representation}. Recent contrastive learning is combined with knowledge distillation, and these contrastive-based distillation approaches~\cite{tian2019crd,xu2020knowledge,zhu2021complementary} aim to capture the correlations and higher-order output dependencies for each sample. Typically, contrastive-based distillation approaches can facilitate cross-resolution knowledge transfer, since they essentially preserve the inter-sample relations which usually are more valuable than the sample representations themselves, especially in visual recognition tasks. The key is the relation modeling for effective knowledge transfer. 

\begin{figure*}[htbp]
\centering
\includegraphics[width=0.92\linewidth]{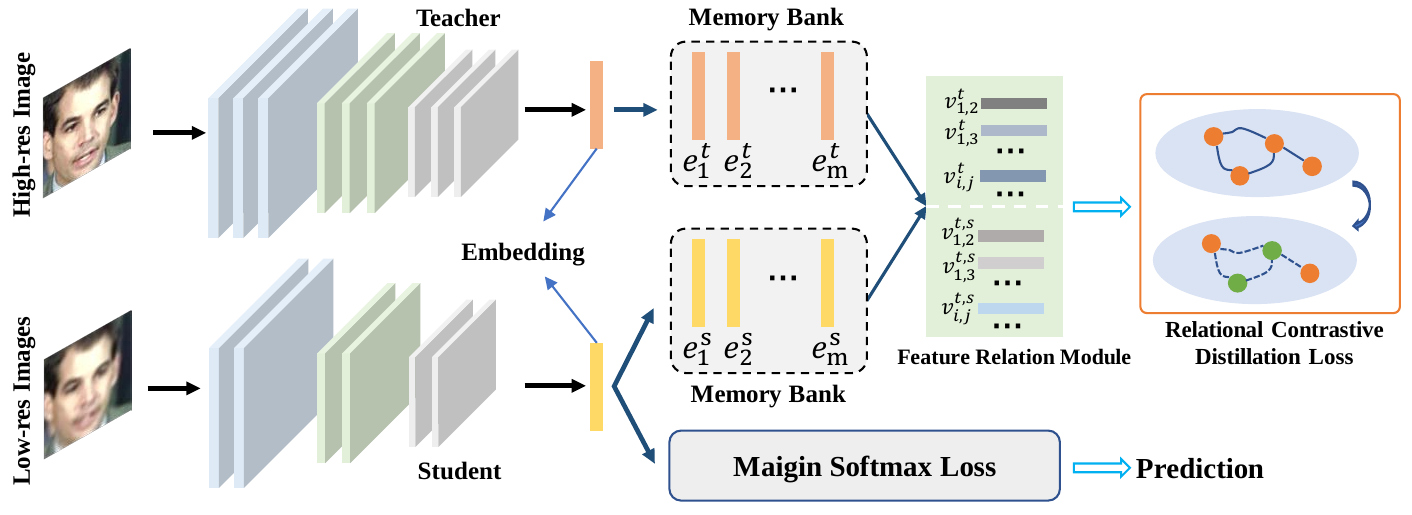}
\caption{The framework of our approach. The approach performs knowledge transfer from high-resolution teacher to low-resolution student by sufficiently modeling high-order representation relations, which simultaneously addresses knowledge distillation and low-resolution recognition in a single framework.}
\label{fig:framework}
\end{figure*}

To transfer high-order dependency within the representation in both relation estimation and knowledge distillation, we propose a teacher-student learning approach for low-resolution object recognition via cross-resolution relational contrastive knowledge distillation with two streams, as shown in Fig.~\ref{fig:framework}. The teacher stream is initialized with a complex pretrained model for high-resolution recognition and the student stream trains a compact model with the help of structural relational knowledge between different resolution samples. By making the high-order relation between low-resolution samples and other high-resolution samples mimic the high-order relation between corresponding high-resolution sample and other high-resolution samples, the student can pay more attention on semantic information instead of pixel details, and then learn the distinction between low-resolution images in the feature space to improve low-resolution object recognition.

Our main contributions are three folds: 1) we propose a cross-resolution relational contrastive distillation approach that is able to distill richer structural knowledge from pretrained high-resolution models to facilitate low-resolution object recognition, 2) we propose a relational contrastive module to extract relational knowledge in contrastive representation space, and 3) we conduct extensive experiments to show the state-of-the-art performance and good adaptability of our approach in low-resolution object recognition.

\section{Related Works}
\subsection{Low-Resolution Object Recognition}
The recognition of low-resolution visual objects is attracting increasing interest due to its widespread applications in long distance surveillance scenarios \cite{zou2011very,li2019low,munir2021resolution}, blurry image analysis~\cite{mahapatra2019image,rouhsedaghat2021low}. Its major challenge is that the informative identity details of the identified objects are seriously missing. In particular, low-resolution objects have less high variance information and the textures can be visually indistinguishable. Recently, an effective way to address this problem is to utilize high-resolution object information for learning improved recognition models. Existing approaches can be categorized into reconstruction-based and prediction-based category. Reconstruction-based approaches employ super-resolution methods to the low-resolution objects before recognition. Grm \etal~\cite{grm2019face} proposed a cascaded super-resolution network, along with an ensemble of face recognition models as identity priors. Chan \etal~\cite{9578478} obtained the effective super-resolution by using the rich and diverse prior knowledge in the pretrained GAN. Kong~\etal~\cite{kong2019cross} proposed resolution invariant model (RIM) to  recognize low-resolution faces from CCTV cameras at different resolutions. RIM uses a tri-path GAN to jointly learn face hallucination sub-net and heterogeneous recognition sub-net. Unfortunately, such approaches require additional computation and the recovered details may be not always beneficial to recognition.

By contrast, prediction-based approaches directly recognize low-resolution objects by knowledge transfer and it is essential to sufficiently represent the domain knowledge and transfer them effectively. On the one hand, a direct approach is transferring the knowledge from high-resolution objects, in which the feature vector distance matters. Soma \etal~\cite{biswas2011multidimensional} proposed to map the low-resolution images to Euclidean space, and then approximate the corresponding high-resolution ones through the distance dimension. Zangeneh \etal~\cite{zangeneh2020low} proposed a new coupled mapping method consisting of two DCNN branches for mapping high and low-resolution face images to non-linear transformed public space. Zha \etal \cite{zha2019tcn} proposed an end-to-end transferable coupling network in high-resolution and low-resolution domains respectively, and introduced a transferable triple loss to narrow cross-resolution positive pairs and separate negative pairs, which improves the recognition performance for low-resolution objects. 

It has been proved feasible using teacher-student learning to transfer knowledge for facilitating visual applications \cite{Hinton2015DistillingTK,gou2021knowledge,liu2022tcsvt}. Such knowledge distillation approaches are mainly based on response, feature and relation. Response-based distillation approaches~\cite{Hinton2015DistillingTK,kim2017transferring,mirzadeh2020improved, zhao2022dkd} aim to directly imitate the neural response of the last output layer of the teacher model. While feature-based distillation approaches~\cite{RomeroBKCGB14,heo2019comprehensive,chen2021cross} mimic the intermediate representations of teacher model to improve the learning of student model by matching original or transformed features. Huang \etal~\cite{huang2022feature} proposed to transfer rich privilege information from a wide and complicated teacher network to a thin and simplified student one. Unlike the above two types of approaches using sample-level outputs of specific layers, relation-level approaches~\cite{tung2019similarity,peng2019correlation,park2019relational,tian2019crd,zhu2021complementary,chen2021cvpr} further explore the relation between data samples, and have shown that transfer structural similarity between instances rather than individual instance representations is beneficial for student learning. Since semantically similar inputs produce similar activations, Tung \etal~\cite{tung2019similarity} used pairwise activation similarities in each input mini-batch to supervise the student learning, and Park \etal~\cite{park2019relational} proposed to transfer explicit sample relations from pretrained teacher. In general, these approaches base on response or low-order relations between samples are often insufficient for cross-resolution knowledge transfer. To address that, we propose a teacher-student learning approach to facilitate low-resolution object recognition via cross-resolution relational contrastive distillation.

\subsection{Contrastive Learning}
Contrastive learning is regarded as a very important part of self-supervised learning, which builds representations by learning to encode what makes two things similar or different. Recent works~\cite{oord2018representation,he2020momentum,chen2020simple} have been widely used to learn the feature representations of samples by comparing the data with positive and negative samples in the feature space. Contrastive losses such as NCE~\cite{hjelm2018learning} and infoNCE~\cite{oord2018representation} measure the similarities of data samples in a deep representation space, which learn representations by contrasting positive and negative representation pairs. One of the major difficulties in contrastive learning is how to construct the positive and negative samples. Deep InfoMAX~\cite{hjelm2018learning} takes local features of training images and different images as positive and negative samples respectively. Instance Discrimination~\cite{wu2018unsupervised} learns to contrast the current embedding with previous embeddings from an online memory bank. The MOCO\cite{he2020momentum} and SimCLR~\cite{chen2020simple} apply augmentation to train samples and requires the network to match original image and transformed images through contrastive loss. These methods only need to learn in the feature space, thus avoiding focus too much on pixel details but paying more abstract semantic information instead.

For knowledge distillation, Tian \etal~\cite{tian2019crd} proposed to combine contrastive learning with knowledge distillation, and Xu \etal~\cite{xu2020knowledge} represented contrastive task as a self-supervised pretext task to facilitate the extraction of richer knowledge from the teacher to the student. They show that incorporating contrastive learning loss into knowledge distillation can help student learn higher-order structural knowledge which can promote cross domain knowledge transfer. They are based on samples and the mutual relations are still insufficient. Thus, it is necessary to explore more effective forms to model the mutual relations of deep representations instead of the representations themselves. Zheng \etal~\cite{zheng2021boosting} proposed relation knowledge distillation by linking cluster-based and contrastive-based self-supervised learning. However, such methods often suffer from poor generalization. To address that, we take into account higher-order relational information between the samples across different image resolutions.

\section{The Approach}
The objective of our cross-resolution relational contrastive distillation (\textbf{CRRCD}) is sufficiently distilling high-order relational knowledge from a pretrained teacher for high-resolution recognition and effectively transferring it to learn a compact student for low-resolution recognition. Toward this end, we build the training instances by taking massive pairs of high-resolution images and corresponding low-resolution images in a self-supervised manner, and utilize vectors to define the representation relations. A feature relation module is utilized to estimate the teacher relation vector in teacher space and the student relation vector in cross-resolution space, respectively. The module is a simple learnable network that consists of two linear layers and a nonlinear activation layer. It is employed to estimate the relation vector between sample representations. Additionally, the cross-resolution relation vector is supervised by its corresponding vector in teacher space. In this manner, relation estimation and representation learning is performed in a unified way. In general, the student is trained on the images from source domain but deployed in target domain, and these two domains often exist large representation discrepancy. Therefore, our relation modeling manner needs to address cross-resolution knowledge transfer with good adaptability.

\subsection{Problem Formulation}
We denote the training set as $\mc{D}=\{(\bm{x}^h_i, \bm{x}^l_i, {y}_i)\}_{i=1}^{|\mc{D}|}$, where $\bm{x}^h_i$ represents the $i$th high-resolution sample with class label ${y}_i\in\{1,2,...,c\}$ and $\bm{x}^l_i$ corresponds to the corresponding low-resolution sample. Here $c$ is the number of classes. Given a teacher network $\phi^{t}$ with parameters $\mathcal{W}^{t}$ and a student network $\phi^{s}$ with parameters $\mathcal{W}^{s}$, we denote the representation of a sample pair $(\bm{x}^h, \bm{x}^l)$ produced by the two networks as $\bm{e}^t=\phi^{t}(\mathcal{W}^{t};\bm{x}^h)$ and $\bm{e}^s=\phi^{s}(\mathcal{W}^{s};\bm{x}^l)$, respectively. Let $(\bm{x}^h_i,\bm{x}^l_i)$ and $(\bm{x}^h_j,\bm{x}^l_j)$ be two sample pairs randomly chosen from the training set. The relation between $\bm{x}^h_i$ and $\bm{x}^h_j$ in teacher space can be modeled as $\bm{v}^{t}_{i,j}$, where $\bm{v}^{t}_i$ is a relation vector produced by the feature relation module $\mathbb{F}$ that takes $\bm{e}^t_i$ and $\bm{e}^t_j$ as inputs. Similarly, we denote $\bm{v}^{t,s}_{i,j}$ as the relation vector across the teacher and student space, the inputs of feature relation module are $\bm{e}^t_i$ and $\bm{e}^s_j$, respectively. The specific form is $v^{t,s} = \varphi\left(\sigma\left(\varphi_{i} \phi^{t}\left(x_{i}\right)-\varphi_{j} \phi^{s}\left(x_{j}\right)\right)\right)$, where $\varphi$ and $\tau$ denote the linear transformation and the ReLU function, respectively. We hope that the cross-space relation $\bm{v}^{t,s}_{i,j}$ can be consistent with $\bm{v}^{t}_{i,j}$ with the help of relational contrastive distillation loss. 

\subsection{Cross-Resolution Relational Contrastive Distillation}
Let $\bm{x}$ represent the input, we denote its empirical data distribution as $p(\bm{x})$. For the conditional marginal distributions $p(\bm{v}^t|\bm{x})$, $p(\bm{v}^{t,s}|\bm{x})$, the sampling procedure is described as:
\begin{equation}
\begin{gathered}
\bm{x}_{i}^{h}, \bm{x}_{j}^{h}, \bm{x}_{i}^{l}, \bm{x}_{j}^{l} \sim p(\bm{x}) \\
\bm{v}_{i, j}^{t}=F^{t}\left(\phi^{t}\left(\mathcal{W}^{t};\bm{x}_{i}^{h}\right), \phi^{t}\left(\mathcal{W}^{t};\bm{x}_{j}^{h}\right)\right) \\
\bm{v}_{i, j}^{t, s}=F^{t, s}\left(\phi^{t}\left(\mathcal{W}^{t};\bm{x}_{i}^{h}\right), \phi^{s}\left(\mathcal{W}^{s};\bm{x}_{j}^{l}\right)\right),
\end{gathered}
\end{equation}
where $F^{t}$ and $F^{t,s}$ are two learnable networks for computing the relation vectors. $\bm{v}_{i, j}^{t}$ and $\bm{v}_{i, j}^{t, s}$ represent the relationship between the $i$-th and $j$-th samples in teacher space and cross-resolution space, respectively. Intuitively, by maximizing Kullback-Leibler (KL) divergence between the joint distribution $p(\bm{v}^{t},\bm{v}^{t,s}|\bm{x})$ and the product of marginal distributions $p(\bm{v}^{t}|\bm{x})p(\bm{v}^{t,s}|\bm{x})$, we can maximize the mutual information (MI) $\mathbb{I}$ between student and teacher representations~\cite{tian2019crd}:
\begin{equation}
\label{eq4}
\mathbb{I}(\bm{v}^{t},\bm{v}^{t,s})=\mathbb{E}_{p(\bm{v}^{t},\bm{v}^{t,s}|\bm{x})} \log \frac{p(\bm{v}^{t},\bm{v}^{t,s}|\bm{x})} {p(\bm{v}^{t}|\bm{x}) p(\bm{v}^{t,s}|\bm{x})}.
\end{equation}

\myPara{MI lower bound.}~To setup an appropriate loss to maximize the mutual information, we define a distribution $q$ with latent variable $b$ which indicates whether the relation tuple ($\bm{v}_{i,j}^t$, $\bm{v}_{i,j}^{t,s}$) is drawn from the joint distribution $(b=1)$ or the product of marginal distributions $(b=0)$:

\begin{equation}
\label{eq5}
\begin{aligned}
&q(\bm{v}^{t},\bm{v}^{t,s} \mid b=1)=p(\bm{v}^{t},\bm{v}^{t,s}) \\
&q(\bm{v}^{t},\bm{v}^{t,s} \mid b=0)=p(\bm{v}^{t}) p(\bm{v}^{t,s}).
\end{aligned}
\end{equation}
Here, $b=1$ means $\bm{v}_{i,j}^t$ and $\bm{v}_{i,j}^{t,s}$ are computed based on the same input pair, and $b=0$ means $\bm{v}_{i,j}^t$ and $\bm{v}_{i,j}^{t,s}$ are independently selected. Now, suppose in our data, we give 1 relevant relation pair ($b=1$) with $n$ irrelevant relation pairs ($b=0$). Then the priors on the latent $b$ are $q(b=1)=1/(n+1)$ and $q(b=0)=n/(n+1)$. By combining the priors with the Bayes’ rule, the posterior for $b=1$ is given by:
\begin{equation}
\label{eq6}
q\left(b=1 \mid \bm{v}^{t},\bm{v}^{t,s}\right)=\frac{p\left(\bm{v}^{t},\bm{v}^{t,s}\right)}{p\left(\bm{v}^{t},\bm{v}^{t,s}\right)+n p\left(\bm{v}^{t}\right) p\left(\bm{v}^{t,s}\right)}.
\end{equation}
Then the mutual information is defined as: 
\begin{equation}
\label{eq7}
\log q\left(b=1 \mid \bm{v}^{t},\bm{v}^{t,s}\right) \leq-\log{n}+\log \frac{p\left(\bm{v}^{t},\bm{v}^{t,s}\right)}{p\left(\bm{v}^{t}\right) p\left(\bm{v}^{t,s}\right)}.
\end{equation}
Taking the expectation on both sides, Eq.~\eqref{eq7} is rewritten as:
\begin{equation}
\label{eq8}
\begin{aligned}
\mathbb{I}(\bm{v}^{t},\bm{v}^{t,s}) & \geq \log{n}+\\
&\mathbb{E}_{q(\bm{v}^{t},\bm{v}^{t,s} \mid b=1)} \log q(b=1 \mid \bm{v}^{t},\bm{v}^{t,s}),
\end{aligned}
\end{equation}
where $\mathbb{I}(\bm{v}^{t},\bm{v}^{t,s})$ is the mutual information between the relation distributions of the teacher and student embedding. Thus maximizing $\mathbb{E}_{q(\bm{v}^{t},\bm{v}^{t,s} \mid b=1)} \log q(b=1 \mid \bm{v}^{t},\bm{v}^{t,s})$ the parameters of the student network will increase a lower bound on mutual information.

\myPara{Relation contrastive loss.} Actually, we maximize the log likelihood of the data under the model to estimate true distribution, which is defined as:

\begin{equation}
\label{eq9}
\begin{aligned}
\mathcal{L}_{critic}(h) &=\mathbb{E}_{q(\bm{v}^{t},\bm{v}^{t,s} \mid b=1)}[\log h(\bm{v}^{t},\bm{v}^{t,s})] \\
& +n \mathbb{E}_{q(\bm{v}^{t},\bm{v}^{t,s} \mid b=0)}[\log (1-h(\bm{v}^{t},\bm{v}^{t,s}))].
\end{aligned}
\end{equation}

\begin{equation}
\label{eq10}
h^{*}=\underset{h}{\arg \max } \mathcal{L}_{\text {critic }}(h) \quad \triangleleft \text { optimal critic }.
\end{equation}
We term $h$ the $critic$ since the representations are learned to optimize the critic’s score.
Considering that the bound in Eq.~\eqref{eq8} and the $\mathbb{E}_{q(\bm{v}^{t},\bm{v}^{t,s} \mid b=1)}[\log h(\bm{v}^{t},\bm{v}^{t,s})]$ is non-positive, we weaken the bound in Eq.~\eqref{eq8},
\begin{equation}
\mathbb{I}(\bm{v}^{t}, \bm{v}^{t,s}) \geq \log{n}+\mathcal{L}_{critic}(h).
\end{equation}
We may choose to represent $h$ with any family of functions that satisfy $h:\left\{\bm{v}^{t},\bm{v}^{t,s}\right\} \rightarrow[0,1]$. In practice, 

\begin{equation}
\label{eqtau}
h\left(\bm{v}^{t}, \bm{v}^{t, s}\right)=\frac{e^{h_{1}\left(\bm{v}^{t}\right) h_{2}\left(\bm{v}^{t, s}\right) / \tau}}{e^{h_{1}\left(\bm{v}^{t}\right) h_{2}\left(\bm{v}^{t, s}\right) / \tau} + n/m},
\end{equation}
where $n$ is the number of negatives, $m$ is the dataset cardinality and $\tau$ is a temperature for adjusting concentration level. $h_1$ and $h_2$ first perform the linear transformation on relations, then normalize the transformed relations with $l_2$ norm.

In our approach, the inputs for the function $h$ are teacher-space relation $\bm{v}^t$ and cross-space relations $\bm{v}^{t,s}$. We aim to maximize the mutual information, which is equivalent to minimizing the relation contrastive loss $\mathcal{L}_{rcd}$:

\begin{equation}
\begin{aligned}
\mathcal{L}_{rcd} &=-\sum_{q(b=1)} \log h\left(\bm{v}^{t}, \bm{v}^{t, s}\right) \\
&-n\sum_{q(b=0)} \log \left[1-h\left(\bm{v}^{t}, \bm{v}^{t, s}\right)\right],
\end{aligned}
\end{equation}
where $\left\{\left(\bm{v}^{t}, \bm{v}^{t, s}\right) \mid b=1\right\}$ acts as positive pairs while $\left\{\left(\bm{v}^{t}, \bm{v}^{t, s}\right) \mid b=0\right\}$ acts as negative pairs.

To achieve superior performance and conduct fair comparisons, we also incorporate the naive knowledge distillation loss $\mathcal{L}_{kd}$ along with our relation contrastive loss. Given the presoftmax logits $\bm{z}^t$ for teacher and $\bm{z}^s$ for student, the naive knowledge distillation loss can be expressed as

\begin{equation}
\mathcal{L}_{k d}=\rho^{2} \mathcal{H}\left(\sigma\left(\bm{z}^{t} / \rho\right), \sigma\left(\bm{z}^{s} / \rho\right)\right),
\end{equation}
where $\rho$ is the temperature, $\mathcal{H}$ refers to the cross-entropy
and $\sigma$ is softmax function. The complete objective is:
\begin{equation}
\mathcal{L}=\mathcal{L}_{c l s}+\alpha \mathcal{L}_{kd}+\beta \mathcal{L}_{rcd},
\end{equation}
where $\mathcal{L}_{cls}$ represents the arcface loss for face recognition, or cross-entropy loss for object classification. We experimentally determine a best combination of the three loss terms, and set $\alpha=0.5$ and $\beta=2$ in our approach.

\myPara{Relationships to similar distillation approaches.}~Like CRD~\cite{tian2019crd} and CRCD~\cite{zhu2021complementary}, our CRRCD is also based on contrastive learning and has a certain similarity in analysis such as a lower bound on the mutual information. Different from them, our approach is designed for cross-quality knowledge transfer in low-resolution recognition task, and the modeling granularity of relational knowledge between samples is finer and the order is higher. Specifically, compared with CRD, CRRCD takes into account higher-order information between samples in different resolution data and requires less negative samples for training. The main differences from CRCD include: 1) CRRCD focuses on the relation between sample representations, while CRCD calculates the relation between sample gradients which may affect the performance of student model detrimentally on low-resolution recognition and increase the cost, 2) CRRCD facilitates cross-resolution knowledge transfer by modeling the relation between samples in different resolution data, while CRCD only transfers information from the same data resolution, 3) CRRCD uses a more efficient critic function Eq.~\eqref{eqtau} to estimate the distribution $q\left(b=1 \mid \bm{v}^{t},\bm{v}^{t,s}\right)$, which helps to maximize a lower bound on the mutual information. Therefore, our CRRCD can achieve better performance on low-resolution object recognition.

\section{Experiments}
To validate the effectiveness of our cross-resolution relational contrastive distillation approach (\textbf{CRRCD}), we conduct experiments on two representative types of applications: low-resolution object classification and low-resolution face recognition. For the low-resolution object classification experiments, we utilize four benchmark datasets: CIFAR100~\cite{krizhevsky2009learning}, SVHN~\cite{netzer2011reading}, STL10~\cite{Coates2011aistats} and TinyImageNet~\cite{TinyImageNet}. The purpose is to assess the performance and generalizability of our approach. Furthermore, we investigate low-resolution face recognition by training models on CASIA-WebFace~\cite{yi2014learning} and evaluating them on three face recognition tasks: verification on LFW~\cite{LFWTech}, identification on UCCS~\cite{uccs} and retrieval on TinyFace~\cite{cheng2018accv}. In these experiments, we employe VGG~\cite{SimonyanZ14a}, ResNet~\cite{HeZRS16}, wide ResNet~\cite{ZagoruykoK16}, ShuffleNetV1~\cite{zhang2018shufflenet} and ShuffleNetV2~\cite{ma2018shufflenet} as as our backbone models. In the model learning process, we use a batch size of 96 and initialize the learning rate to 0.05. The learning rate is multiplied by 0.1 at epochs 21, 28, and 32. We maintain a fixed random seed of 5 and set the distillation temperature (T) to 4. All experiments are conducted with PyTorch on a NVIDIA 3090 GPU.

\subsection{Low-resolution Object Classification}
Object classification is a general visual recognition task and has very important applications under the low-resolution condition like industrial inspection and medical diagnosis. In the experiments, we first check the effectiveness of our distillation method and then evaluate the effectiveness and transferability of our approach in low-resolution object classification. 

\begin{table*}[htbp]
	\centering
	\caption{Classification accuracy (\%) with peer-architecture setting on CIFAR100. The best and the second best results are in \textbf{bolded} and \underline{underlined}, respectively.}
	\begin{tabular}{l|ccccccc}
		\hline
		Teacher Network & WRN-40-2 & WRN-40-2  & ResNet56 & ResNet110 & ResNet110 & ResNet32x4  & VGG13 \\
		Student Network & WRN-16-2  & WRN-40-1 & ResNet20 & ResNet20  & ResNet32 & ResNet32  & VGG8 \\
		\hline
		Teacher  & 75.61 & 75.61  & 72.34  & 74.31 & 74.31 & 79.42 & 74.64   \\
		Student  & 73.26 & 71.98  & 69.06  & 69.06 & 71.14 & 72.50 & 70.36   \\
		\hline
		KD~\cite{Hinton2015DistillingTK} & 74.92 & 73.54 & 70.66 & 70.67  & 73.08 & 73.33 & 72.98 \\
		Fitnet~\cite{RomeroBKCGB14}  & 73.58  & 72.24 & 69.21 & 68.99 & 71.06 & 73.50 & 71.02 \\
		AT~\cite{ZagoruykoK17} & 74.08  & 72.77 & 70.55 &  70.22 & 72.31 & 73.44 & 71.43 \\
		PKT~\cite{pkt_eccv}  & 74.54 &  73.45 & 70.34 & 70.25 & 72.61 & 73.64  & 72.88 \\
		SP~\cite{tung2019similarity} & 73.83 & 72.43 & 69.67 & 70.04 & 72.69 & 72.94  & 72.68 \\
		RKD~\cite{park2019relational} & 73.35 & 72.22 & 69.61 & 69.25 & 71.82 & 71.90 & 71.48 \\
		CC~\cite{peng2019correlation} & 73.56  & 72.21 & 69.63 & 69.48 & 71.48 & 72.97  & 70.71 \\
		VID~\cite{ahn2019variational} 	& 74.11 & 73.30 & 70.38 & 70.16 & 72.61 & 73.09 & 71.23 \\
		Abound~\cite{heo2019knowledge}  & 72.50 & 72.38 & 69.47 & 69.53 & 70.98 & 73.17 & 70.94 \\
		CRD~\cite{tian2019crd} & 75.48 & {74.14} & 71.16 & 71.46 & 73.48 & {75.51} & 73.94 \\
		CRCD~\cite{zhu2021complementary} & \underline{76.37} & 73.84 & 70.89 &  70.98 & 73.32 & 73.50 & 73.89 \\
		WCoRD~\cite{chen2021cvpr} & 75.88 & {74.73} & {71.56} & \underline{71.57} & {73.81} & \underline{75.95}  & \underline{74.55} \\
		DKD~\cite{zhao2022dkd} & 76.24 & \underline{74.81} & \underline{71.97} & -- & \textbf{74.11} & \textbf{76.32} & \textbf{74.68} \\
		\textbf{CRRCD}&  \textbf{76.43}  & \textbf{74.83} & \textbf{72.10} &  \textbf{71.92}  & \underline{74.05} & 75.14 & {74.04} \\
		\hline
	\end{tabular}
	\label{tab4}
\end{table*}

\begin{table*}[htbp]
	\centering
	\caption{Classification accuracy (\%) with cross-architecture setting on CIFAR100 . The best and the second best results are in \textbf{bolded} and \underline{underlined}, respectively.}
	\begin{tabular}{l|ccccc}
		\hline
		Teacher Network & ResNet18 & VGG11 &~ResNet18 & WRN50-2 & WRN50-2\\
		Student Network & VGG8 & ShuffleNetV1 & ShuffleNetV2 & ShuffleNetV1 &~VGG8 \\
		\hline
		Teacher & 76.61 & 70.76 & 76.61 & 80.24 &80.24\\
		Student & 69.21 & 66.18 & 70.48 & 66.18 & 69.21\\
		\hline
		Factor~\cite{KimPK18} & 68.06 & 68.16 & 69.99 & 70.51 & 70.12\\
		KD~\cite{Hinton2015DistillingTK} & 71.17 & 72.40 & 75.03 & 71.78 & 70.31\\
		Fitnet~\cite{RomeroBKCGB14} & 70.59 & 70.50 & 72.24 & 70.46 &70.04\\
		AT~\cite{ZagoruykoK17} & 71.62 & 69.64 & 73.83 & 70.55 & 69.78 \\
		PKT~\cite{pkt_eccv} & 72.74 & 72.06 & 74.31 & 69.80 &69.76\\
		RKD~\cite{park2019relational} & 71.03 & 7 0.92 & 73.26 & 70.58 & 70.41\\
		SP~\cite{tung2019similarity} & 73.07 & 72.31 & 74.95 & 70.70 & 70.00\\
		CC~\cite{peng2019correlation} & 69.82 & 70.70 & 72.21 & 70.66 & 69.96 \\
		VID~\cite{ahn2019variational} & 71.75 & 70.59 & 72.07 & 71.61 & 71.00 \\
		Abound~\cite{heo2019knowledge} & 70.42 & 72.56 & 74.64 &~\underline{74.28} & 69.81 \\
		CRD~\cite{tian2019crd} & 73.17 & 72.38 & 74.88 & 71.08 & 72.50 \\
		CRCD~\cite{zhu2021complementary} & {73.54}& {73.07} & \underline{75.35} & 73.58 & \underline{74.13}\\
		EKD~\cite{zhang2021student} & \underline{73.82}& \underline{73.18} & {75.26} & 73.61 & {74.05}\\
		\textbf{CRRCD} & \textbf{74.49}& \textbf{74.35} & \textbf{77.06} & \textbf{76.53} & \textbf{74.26}\\
		\hline
	\end{tabular}
	\label{tab-cross}
\end{table*}
 
\myPara{The effectiveness of distillation.}~Our approach distills cross-resolution contrastive relations between different resolution samples that can better mimic the model capacity of the high-resolution teacher model. To verify that, we conduct two low-resolution object classification experiments on CIFAR100 by comparing with other advanced distillation approaches under both peer-architecture and cross-architecture settings. CIFAR100 has 100 classes containing 600 images each.

\textbf{Peer-architecture distillation} uses homogeneous architecture for teacher-student pairs. The results are shown in Tab.~\ref{tab4}. From the results, we can see that our CRRCD outperforms six sample-level distillation approaches (KD~\cite{Hinton2015DistillingTK}, FitNet~\cite{RomeroBKCGB14}, AT~\cite{ZagoruykoK17}, PKT~\cite{pkt_eccv}, VID~\cite{ahn2019variational} and Abound~\cite{heo2019knowledge}) as well as six relation-level distillation approaches (SP~\cite{tung2019similarity}, RKD~\cite{park2019relational}, CC~\cite{peng2019correlation}, CRD~\cite{tian2019crd}, CRCD~\cite{zhu2021complementary} and WCoRD~\cite{chen2021cvpr}), and is comparable with DKD~\cite{zhao2022dkd}. For example, comparing with WCoRD~\cite{chen2021cvpr} that combines contrastive learning and knowledge distillation to help student learn richer sample-wise knowledge in a certain maturity, when taking ResNet56 as teacher and ResNet20 as student, our CRRCD achieves 72.10\% accuracy on CIFAR100 which is 0.54\% higher than WCoRD, and gains 0.24\% improvement when the teacher and student is ResNet110 and ResNet32. The main reason comes from that our CRRCD focuses on higher-order relational contrasting knowledge. It implies the remarkable effectiveness in improving student learning.

To further explore the flexibility of our approach, \textbf{cross-architecture distillation} applies heterogeneous architecture for teacher-student pairs during learning. In this setting, the gap of knowledge transfer will become larger thus put forward higher requirements for knowledge distillation. The results are shown in Tab.~\ref{tab-cross}, where our approach achieves the best accuracy and has better competitiveness than peer-architecture setting. For five cross-architecture students, our CRRCD gains 2.54\% improvement over CRD and 1.40\% improvement over CRCD on average accuracy, respectively. Especially, when taking WRN50-2 as teacher and ShuffleNetV1 as student, CRRCD achieves 5.45\% accuracy improvement over CRD and 1.97\% accuracy improvement over CRCD, respectively. Moreover, compared to recent evolutionary knowledge distillation approach (EKD)~\cite{zhang2021student}, our CRRCD also gives better classification accuracy. These results shows that our approach can provide a flexible way to distill black-box teacher knowledge and learn discriminative student representations for downstream image recognition task. 

\myPara{Very low-resolution object classification.}~First, we check the effectiveness of CRRCD on object classification under a very low-resolution of $8 \times 8$ by evaluating on SVHN dataset. This dataset contains digit images captured from real-world natural scenes, having a resolution of $32\times32$. We downsample the images by a factor of 4 to create $8\times8$ data and use them for evaluating very low-resolution digit classification. The teacher model is ResNet56 pretrained with $32\times32$ images and our student is VGG8 that has very few parameters. We compare our approach to five state-of-the-art very low-resolution image recognition approaches and report the top-1 classification accuracy in Tab.~\ref{tab:svhn}. Our CRRCD model obtains a classification accuracy of 89.33\%, an at least improvement of 1.58\%. Compared with other approaches like DeriveNet~\cite{singh2021derivenet} which focuses on learning effective class boundaries by utilizing the class-specific domain knowledge, our CRRCD makes full use of the structural knowledge between different samples and the dark knowledge in the teacher model to obtain stronger feature extraction capability, which greatly improves the recognition performance of model on very low-resolution images.

\begin{table}[t]
  \centering
  
  \caption{Very low-resolution~($8\times 8$) recognition on SVHN.}
    \begin{tabular}{l|c|c}
    	\hline
    Algorithm  & Accuracy (\%)  & Publication\\
    \hline
    RPC Nets~\cite{wang2016studying} & 56.89 & CVPR 2016\\
    SICNN~\cite{zhang2018super} & 81.53 & ECCV 2018\\
    DirectCapsNet~\cite{singh2019dual} & 84.51 & ICCV 2019\\
    CSRIP~\cite{grm2019face} & 84.61 & TIP 2019\\
    DeriveNet~\cite{singh2021derivenet} & 87.85 & TPAMI 2022\\
    \textbf{CRRCD} & \textbf{89.33}  &--\\
    \hline
    \end{tabular}
  \label{tab:svhn}
\end{table}

\begin{table*}[htbp]
	\centering
	\caption{Linear classification accuracy (\%) on STL10 and TinyImageNet}
	\renewcommand{\arraystretch}{1.1}
	\begin{tabular}{c|cc|ccccccc}
		\hline
		 & Student & Teacher & KD~\cite{Hinton2015DistillingTK} & FitNet~\cite{RomeroBKCGB14} & AT~\cite{ZagoruykoK17}  &  CRD~\cite{tian2019crd} &  SSKD~\cite{xu2020knowledge}  & \textbf{CRRCD}\\
		\hline
		CIFAR100$\rightarrow$ STL10  & 71.58 & 71.01 & 73.25 & 73.77 & 73.47 & 74.44 & 74.74 & \textbf{75.15}\\
		CIFAR100$\rightarrow$ TinyImageNet  & 32.43 & 27.74 & 32.05 & 33.28 & 33.75 & 34.30  & 34.54& \textbf{35.17}\\
		\hline
	\end{tabular}
	\label{tinyimagenet}
\end{table*}

\myPara{Representation transferability.}~After the promising results achieved with the adaptability on low resolution and flexible network architectures, we further verify the cross-dataset transferability of our approach by training on CIFAR100 but testing on STL10 and TinyImageNet. Following CRD~\cite{tian2019crd}, we investigate the effectiveness of student representations. A good representation extractor should generate linear separable features. Hence, we use the fixed backbone of student trained on CIFAR100 to extract representations for STL10 and TinyImageNet, and then train a linear classifier to test the classification accuracy. We select WRN-40-2 as teacher and ShuffleNetV1 as student, and compare with three sample-level distillation approaches (KD~\cite{Hinton2015DistillingTK}, FitNet~\cite{RomeroBKCGB14} and AT~\cite{ZagoruykoK17}), relation-level distillation approach CRD~\cite{tian2019crd} and self-supervised knowledge distillation (SSKD)~\cite{xu2020knowledge}. In the experiment, the input resolution of teacher and student is $32 \times 32$. As shown in Tab.~\ref{tinyimagenet}, our CRRCD delivers the best accuracy on both STL10 and TinyImageNet. From the results, we find that our approach still has good representation transferability between different objects (e.g., natural objects in CIFAR100 and digits in STL 10). However, all approaches achieve a very low accuracy (e.g., lower than 36\%) in recognizing TinyImageNet. The main reason may be insufficient knowledge from $32 \times 32$ CIFAR100 that is incapable for identifying higher-resolution objects in TinyImageNet. It implies that direct learning from low-resolution images may be ineffective and cross-resolution knowledge transfer can be a more effective way.

\subsection{Low-resolution Face Recognition}
Low-resolution face recognition is a specified and challenging object recognition task and has very helpful applications like recognizing surveillance faces in the wild. In practical scenarios, the facial images often have low resolution, uneven light intensity, diverse facial posture and facial expression. These will have a huge impact on the recognition accuracy. In our experiments, we take CASIA-WebFace as training set, which contains 10575 categories and a total of 494414 images collected from the web. The teacher is trained on CASIA-WebFace with ResNet50 under the high-resolution of $112\times112$, and the students are trained on low-resolution CASIA-WebFace with ResNet18. Then, the trained students are used to evaluate face verification on LFW, face identification on UCCS and face retrieval on TinyFace, respectively. In order to verify the validity of the low-resolution students, we emphatically check the accuracy when the input resolution is 16$\times$16 produced by bilinear downsampling. All approaches use the same experimental settings to ensure fair comparisons. 

\begin{table}[htbp]
  \centering
  \caption{Face verification performance on LFW. Our student achieves good accuracy at a much low resolution of 16$\times$16.}
    \begin{tabular}{c|c|c|c}
    	\hline
    Model & Resolution & Accuracy(\%) &Publication  \\
    \hline
    DeepFace~\cite{parkhi2015deep} & 152$\times$152  & 97.35 &BMVC 2015\\
    DeepID2~\cite{sun2015deep} & 55$\times$47  & 99.15 &NeurIPS 2015\\
    FaceNet~\cite{schroff2015facenet} & 96$\times$96 & 99.63 &CVPR 2015\\
    MobileID~\cite{luo2016face} & 55$\times$47 & 98.37 &AAAI 2016\\
    SphereFace~\cite{liu2017sphereface} & 112$\times$96 & 99.42 &CVPR 2017\\
    ShiftFace~\cite{wu2018shift} & 224$\times$224  & 96.00   &CVPR 2018 \\
    CosFace~\cite{wang2018cosface}  & 112$\times$96 & 99.73 &CVPR 2018\\
    VGGFace2~\cite{Cao2018VGGFace2} & 224$\times$224  & 99.53 &FG 2018\\
    ArcFace~\cite{deng2019arcface} & 112$\times$112  & 99.82 &CVPR 2019\\
    GroupFace~\cite{Groupface} & 112$\times$112  & 99.85 &CVPR 2020\\
    MagFace~\cite{meng2021magface} & 112$\times$112 & 99.83 &CVPR 2021\\
    \hline
    FaceNet~\cite{schroff2015facenet} & 16$\times$16 & 90.25 &CVPR 2015\\
    CosFace~\cite{wang2018cosface}  & 16$\times$16 & 93.80 &CVPR 2018\\
    ArcFace~\cite{deng2019arcface}  & 16$\times$16 & 92.30 &CVPR 2019\\
    MagFace~\cite{meng2021magface} & 16$\times$16 & 94.97 &CVPR 2021\\
    \hline
    SKD~\cite{ge2018low}   & 16$\times$16 & 85.87  &TIP 2019\\
    HORKD~\cite{GeHORKD20}    & 16$\times$16 & 90.03 &AAAI 2020\\
    NPM~\cite{rouhsedaghat2021low}   & 16$\times$16 & 82.16  &PRL 2021\\
    EKD~\cite{zhang2021student} & 16$\times$16 & 91.71  &TCSVT 2022\\
    RPCL-CosFace~\cite{li2022deep} & 16$\times$16 & 95.13 & NN 2022\\
    RPCL-ArcFace~\cite{li2022deep} & 16$\times$16 & 94.70 & NN 2022\\
    RPCL-MagFace~\cite{li2022deep} & 16$\times$16 & 95.12 & NN 2022\\
    \textbf{CRRCD}    & 16$\times$16 & \textbf{95.25} &--\\
    \hline
    \end{tabular}
  \label{tab:lfw}
\end{table}

\myPara{Face verification on LFW.}~We conduct the comparisons with some state-of-the-art face recognition models on LFW, which contains 6000 pairs of face images. We downsample the images to synthesize low-resolution faces. A 512$d$ feature embedding for each image is extracted for similarity comparison. With a pre-set threshold, each face pair is determined to have the same identity if the similarity of the two faces is greater than the threshold and different identity otherwise. The verification accuracy is reported as the percentage of the pairs that are correctly determined. The results are listed in Tab.~\ref{tab:lfw}, where some conclusions can be found.

Firstly, the state-of-the-art face recognition models usually deliver very high verification accuracy in recognizing faces under normal resolution. For example, ArcFace~\cite{deng2019arcface} uses ResNet50 and gives a 99.82\% accuracy under the input resolution of $112\times112$. Our CRRCD approach distills the ResNet50 model into a lightweight ResNet18 student, which still achieves a good accuracy of 95.25\% under a much low-resolution of $16\times16$. This is very helpful for practical deployment in resource-limited conditions. Secondly, when these face recognition models are applied to identify low-resolution images, e.g., recognizing $16\times16$ images after bilinear upsampling, the accuracy will has a great drop. For example, ArcFace gives an accuracy of 92.30\% under the low-resolution condition, having a drop of 7.52\%. These results reveal that it is necessary to compensate the missing knowledge to facilitate the recognition of low-resolution objects from high-resolution images or models. Finally, we compare our approach with five recent low-resolution face recognition approaches. In comparison to distillation-based methods, our CRRCD achieves higher accuracy. This improvement can be attributed to its ability to extract high-order relation contrastive knowledge, which proves to be more effective than sample-level knowledge (SKD~\cite{ge2018low} and EKD~\cite{zhang2021student}) or low-order relation knowledge (HORKD~\cite{GeHORKD20}). In low-resolution face recognition tasks, our method exhibits significant advantages compared to the non-parametric low-resolution face recognition model (NPM~\cite{rouhsedaghat2021low}). In~\cite{li2022deep}, deep Rival Penalized Competitive Learning (RPCL) is embedded into state-of-the-art face recognition models to learn margin-based discriminative low-resolution face features. Our CRRCD outperforms RPCL-based models since it implicitly encodes margin-based discriminative representation learning by using anchor-based high-order relation preserving distillation. In cross-resolution knowledge transfer, high-order relation can help the model learn better representations from low-resolution domain and contrastive relation can facilitate the learning of representations in visual recognition task.

\begin{table}[t]
	\centering
	\caption{Face identification performance on UCCS under a low resolution of $16\times16$. Our student outperforms 12 approaches by at least an accuracy improvement of 1.46\%.}
	\renewcommand{\arraystretch}{1.1}
	\begin{tabular}{l|c|c}
		\hline
		Model & Accuracy(\%) & Publication\\
		\hline
		VLRR~\cite{wang2016studying}    & 59.03   &CVPR 2016\\
		SphereFace~\cite{liu2017sphereface} & 78.73  &CVPR 2017\\
		CosFace~\cite{wang2018cosface} & 91.83  &CVPR 2018\\
		VGGFace2~\cite{Cao2018VGGFace2}   & 84.56  &FG 2018\\
		ArcFace~\cite{deng2019arcface}  & 88.73  &CVPR 2019\\
		SKD~\cite{ge2018low}   & 67.25  &TIP 2019\\
		AGC-GAN~\cite{talreja2019attribute}   & 70.68 &BTAS 2019\\
		LRFRW~\cite{li2019low}   & 93.40  &TIFS 2019\\
		CSRIP~\cite{grm2019face} & 93.49 &TIP 2019\\
		DirectCapsNet~\cite{singh2019dual} & \underline{95.81} &ICCV 2019\\
		HORKD~\cite{GeHORKD20}   & 92.11  &AAAI 2020\\
		EKD~\cite{zhang2021student} & 93.85  &TCSVT 2022\\
		\textbf{CRRCD} & \textbf{97.27}  &--\\
		\hline
	\end{tabular}
	\label{tab:uccs}
\end{table}

\begin{table}[htbp]
	\centering
	\caption{Face retrieval accuracy on TinyFace.}
	\renewcommand{\arraystretch}{1.1}
	\begin{tabular}{l|ccc|c}
		\hline
		Model & Rank-1 & Rank-10 & Rank-20 & Publication\\
		\hline
		CenterFace~\cite{wen2016discriminative} & 0.32 & - & 0.45 &ECCV 2016\\
		DCR~\cite{lu2018deep} & 0.29 & 0.40 & 0.44  &SPL 2018 \\
		PeiLi's~\cite{li2018face}  & 0.31 & 0.43 & 0.46 &Arxiv 2018\\
		CosFace~\cite{wang2018cosface} & 0.29 & 0.39 & 0.42 &CVPR 2018\\
		ArcFace~\cite{deng2019arcface} & 0.26 & 0.34 & 0.37 &CVPR 2019\\
		MagFace~\cite{meng2021magface} & 0.33 & 0.44 & 0.47 &CVPR 2021\\
		RPCL~\cite{li2022deep} & 0.34 & 0.45 & 0.49  &NN 2022\\
		\textbf{CRRCD} & \textbf{0.35} & \textbf{0.47} & \textbf{0.50} &--\\ 
		\hline
	\end{tabular}
	\label{tab:tinyface}
\end{table} 
\myPara{Face identification on UCCS.}~UCCS is collected in real surveillance scenarios and contains 16149 images in 1732 subjects in the wild condition, which is a very challenging benchmark with various levels of challenges. To verify the robustness of our low-resolution student models, we emphatically check the accuracy when the input resolution is 16$\times$16. We follow the setting as~\cite{wang2016studying,ge2018low}, randomly select a 180-subject subset, separate the images into 3918 training images and 907 testing images, and report the results with the standard accuracy. In the experiment, we freeze the representation extraction part of each model, modify the final softmax layer into 180-way, and finetune the layer parameters on training set. As shown in Tab.~\ref{tab:uccs}, our student model achieves an impressive identification accuracy of 97.27\%, surpassing the state-of-the-art DirectCapsNet~\cite{singh2019dual} by 1.46\%. Our approach enhances low-resolution face recognition performance by enabling the student model to acquire discriminative representations. Despite lacking essential information for recognition, our method leverages cross-resolution relational contrastive knowledge from the teacher model and high-resolution data. This allows the student model to learn higher-order feature representations, leading to improved performance.

\myPara{Face retrieval on TinyFace.}~TinyFace contains large-scale native low-resolution surveillance face images. In experiment, we finetune basic models on its training set and then evaluate 1:N identification performance on its testing set. Tab.~\ref{tab:tinyface} reports Rank-1, Rank-10 and Rank-20 retrieval results. Different from typical models~\cite{wen2016discriminative,wang2018cosface,deng2019arcface,meng2021magface,li2022deep} that design margin-based losses to learn discriminative representations, our CRRCD implicitly learns distinct inter-class boundaries under cross-resolution relational constraints with the assistance of a high-resolution teacher and consistently improves retrieval accuracy under various settings. In addition, via high-order knowledge transfer, CRRCD outperforms PeiLi's method~\cite{li2018face} based on reconstruction and DCR~\cite{lu2018deep} that employs two branches to transfer cross-resolution knowledge by feature approximation. There results imply the effectiveness of our approach in learning discriminative and transferable representations.

\subsection{Ablation and Further Analysis}

\begin{figure}[t]
 \centering
 {\includegraphics[width=1.70in]{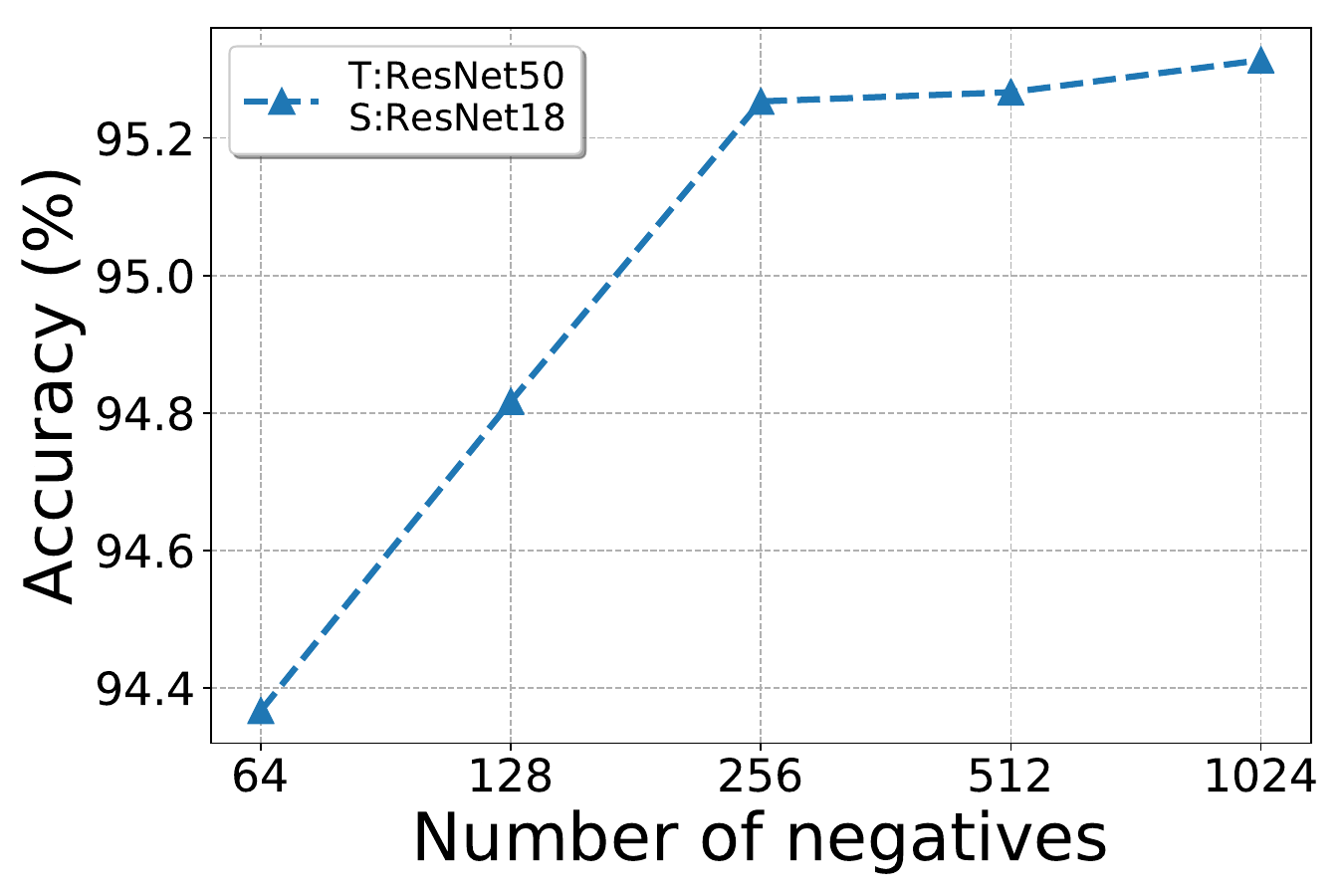}}
 {\includegraphics[width=1.70in]{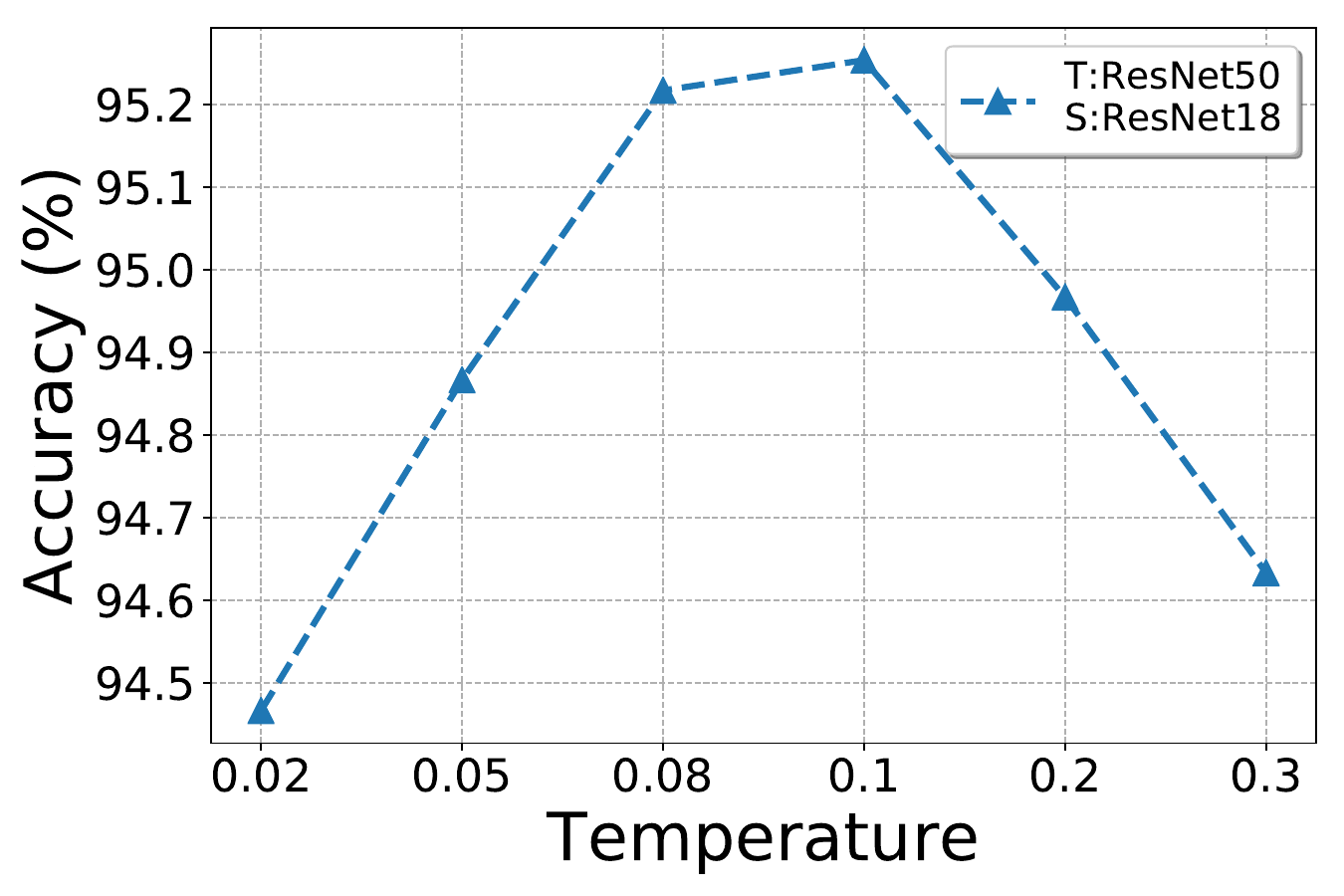}}
 \caption{Face verification accuracy on LFW under different negative number (left) and distillation temperature (right).}
 \label{fig3}
\end{figure}

\myPara{Effect of negative number.} An important part of knowledge distillation based on contrastive learning is to construct positive and negative sample pairs, and the negative number has a crucial impact on the final performance. We validate five different negative number (64, 128, 256, 512 and 1024) and show the results in the left of Fig.~\ref{fig3}. Here, increasing negative number will lead to performance improvement, which means higher-order relation knowledge is built and migrated. Meantime, a larger negative number requires more computations. It suggests that the negative number should be carefully selected to balance the accuracy and computation cost. Thus, we set the negative number to 512 since it only gives a small accuracy gain of 0.05\% when increasing negative number to 1024. Our approach can significantly reduce the negative number, which is benefited from modeling the structural relationship that does not pass through the samples with rich knowledge, which reduces the dependence on the number of negative samples.

\myPara{Effect of sampling policy.} We consider two negative sampling policies when giving an anchor $\bm{x}_i$: $\bm{x}_j,j \neq i$ for the unsupervised case without labels, or $\bm{x}_j, y_j \neq y_i$ for supervised case, where $y_i$ represents the label associated with sample $\bm{x}_i$. What's more to ensure that negative samples are as up-to-date as possible, we store features and gradients in a queue way which will remove the oldest sample when adding the latest sample. Through experiments, the combination of queue and supervised sampling policy can bring at least 0.25\% improvement at accuracy on LFW.

\myPara{Effect of distillation temperature.} The distillation temperature $\tau$ in Eq.\eqref{eqtau} is used to adjust the concentration level. We report the results when $\tau$ varies from 0.02 to 0.30 in the right of Fig.~\ref{fig3}. A temperature between 0.08 to 0.1 works well and we set $\tau=0.1$ for all our experiments. In general, for different downstream tasks, the value of $\tau$ should be carefully set in a task-specified manner.

\myPara{Effect of projected feature dimension.} Our feature relation module builds contrastive relation vectors by projecting the 512$d$ feature embeddings into specific-dimensional features. The projected feature dimension affects model performance and computation cost in training. Increasing dimension boosts performance but also raises computation cost. To balance them, we test various feature dimensions and set it to 128. In addition, our approach employs an efficient critic function $h\left(\bm{v}^{t},\bm{v}^{t,s}\right)$ to estimate the distribution $q\left(b=1 \mid \bm{v}^{t},\bm{v}^{t,s}\right)$, which maximizes a lower bound on the mutual information. It is worth noting that the inference complexity is fixed and not affected by the order of structural relationship.

\begin{figure}[htbp]
    \centering 
    \includegraphics[width=0.48\linewidth]{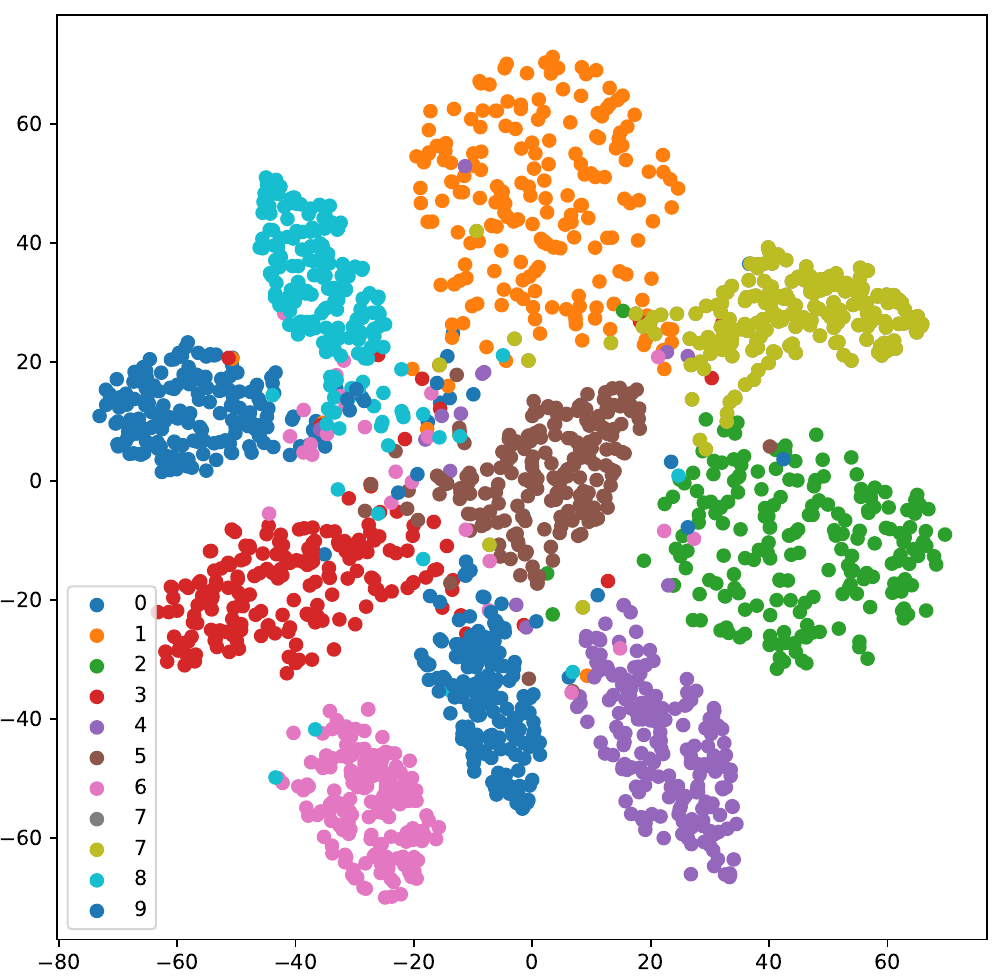}
    \includegraphics[width=0.48\linewidth]{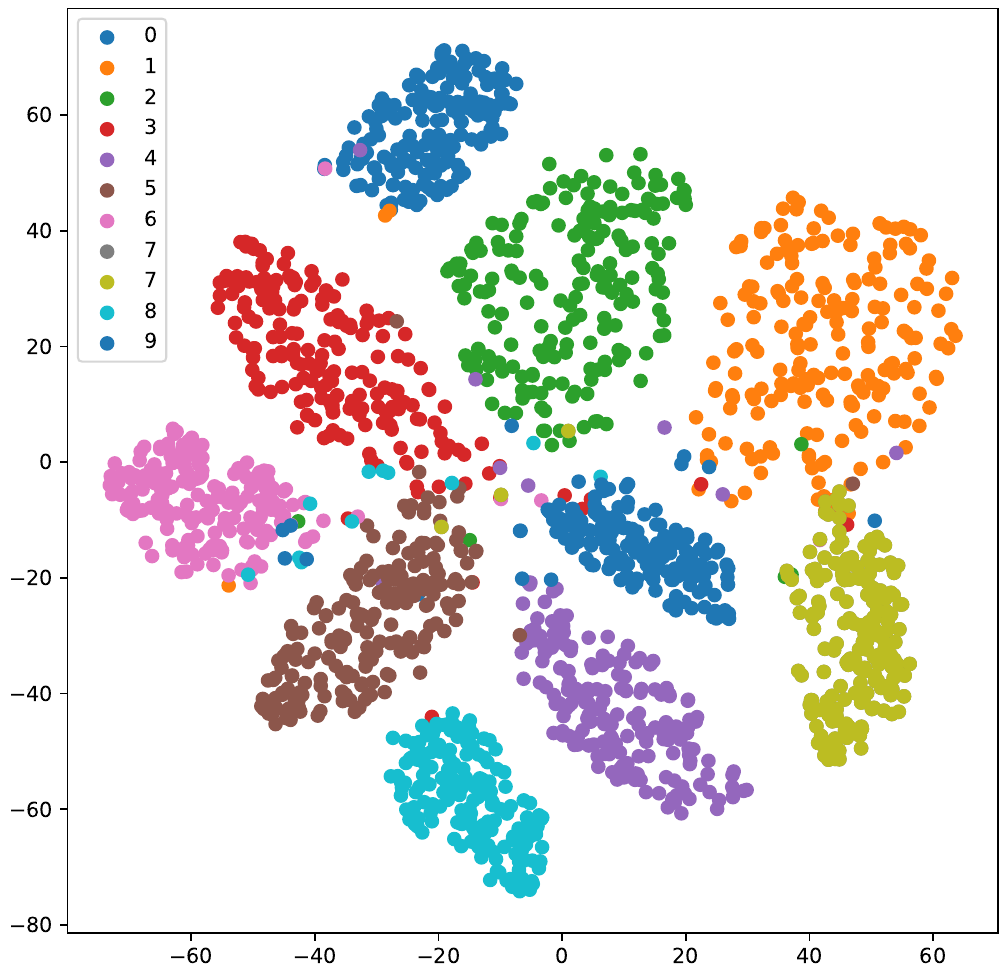}
    \caption{t-SNE feature plots by baseline (left) trained with softmax loss and CRRCD (right) on SVHN.}
    \label{fig:tsne}
\end{figure}

\begin{figure}[htbp]
 \centering
 {\includegraphics[width=0.48\linewidth]{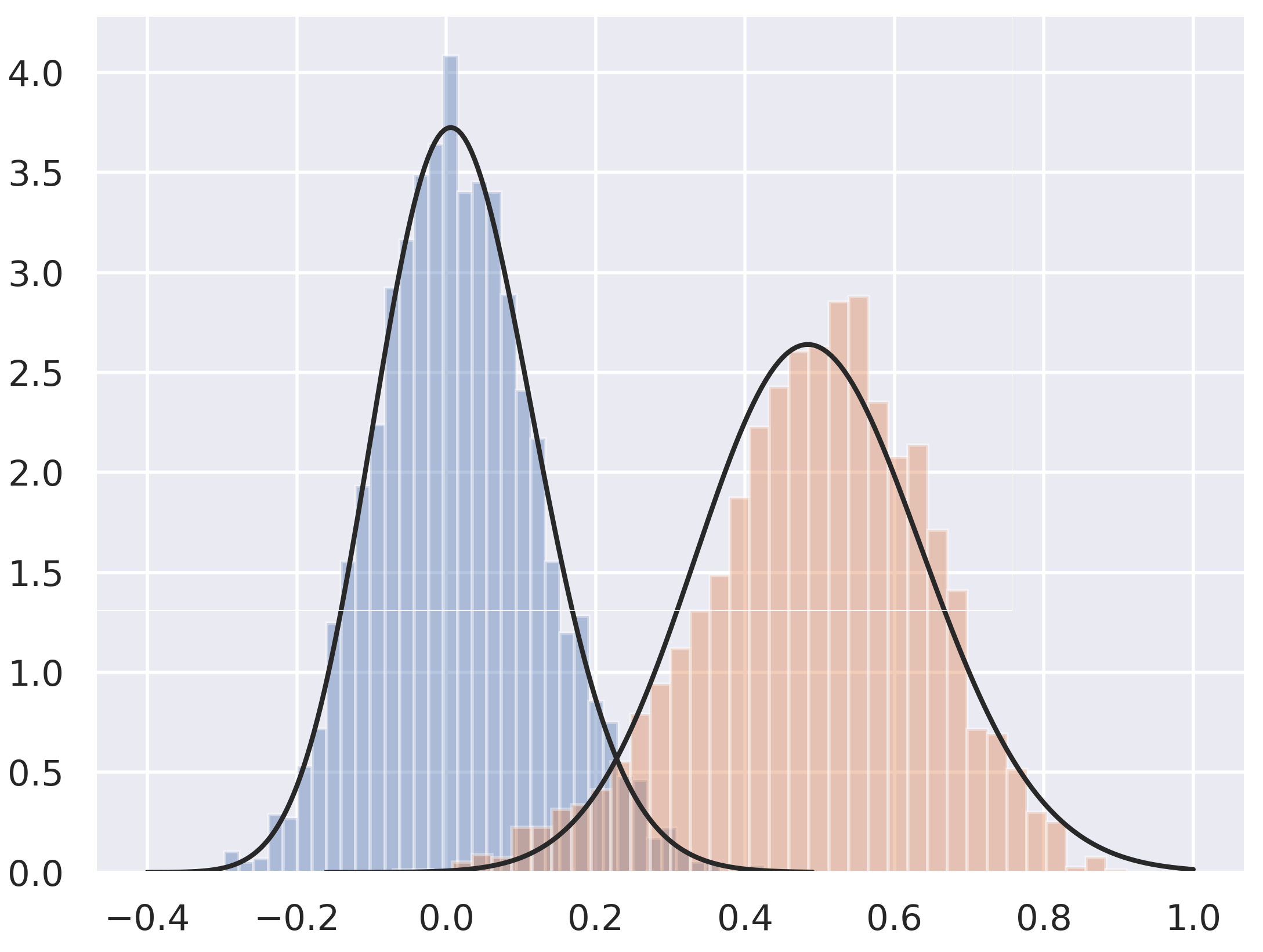}}
 {\includegraphics[width=0.48\linewidth]{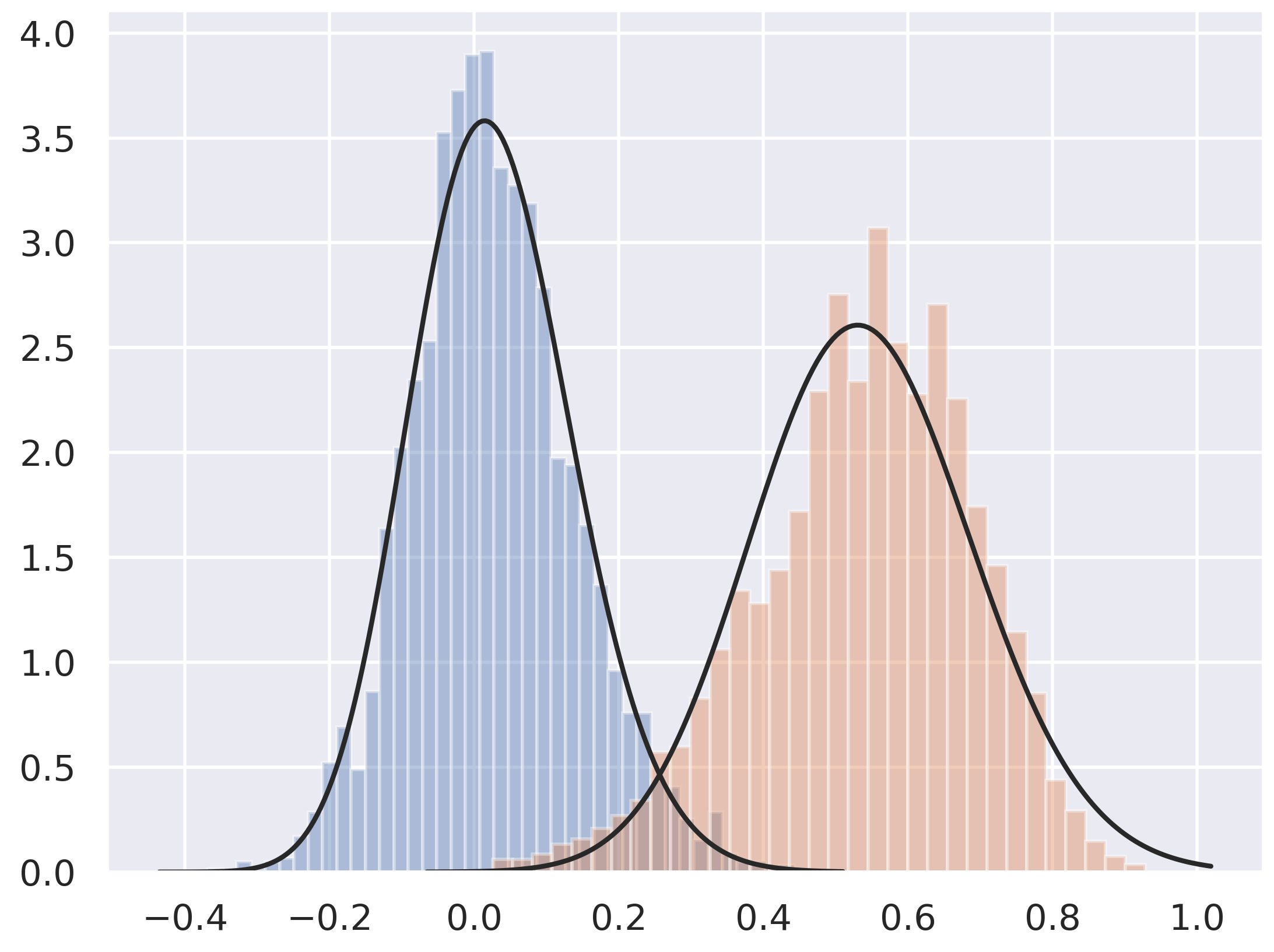}}
 \caption{The distribution of cosine similarity score under low-resolution setting on LFW by ArcFace (left) and CRRCD (right). The x-axis represents the cosine similarity of face pairs, and y-axis is the frequency. The negative pairs and positive pairs are marked in blue and orange, respectively.
}
 \label{fig_d}
\end{figure}

\myPara{Representation visualization.} To further demonstrate the advantages of our approach visually, we first use the t-SNE~\cite{van2008visualizing} for visualization. It converts similarities between data points to joint probabilities and tries to minimize the KL divergence between the joint probabilities of the low-dimensional embedding and the high-dimensional data. We randomly select 400 samples each class from SVHN dataset, different numbers indicate different classes in Fig.~\ref{fig:tsne}. It is obvious that our approach achieves more concentrated clusters than baseline (Same structure as student model, but no distillation strategy) which is trained with softmax loss. And the changes of the distances in classifiers of baseline are more severe than that in classifier of CRRCD. We speculate that transferring high-order relational contrastive knowledge is helpful for student to learn discriminative representations.

Next, we illustrate the estimated similarity distributions of ArcFace and our CRRCD in Fig.~\ref{fig_d}. To quantify their difference, we introduce two statistics for evaluation, the expectation margin and histogram intersection between the two distributions from positive and negative pairs. Typically, smaller histogram
intersection and larger expectation margin indicate better verification performance, since it means that more discriminative deep embeddings are learned. As shown in Fig.~\ref{fig_d}, the deeply learned face features are more discriminative and less overlapped by our CRRCD than by ArcFace, indicating that our approach is effective in enhancing the discriminability and obtains the best performance.

\section{Conclusion}
In this paper, we propose cross-resolution relational contrastive distillation, a novel approach to improve low-resolution object recognition. Our approach successfully transfers high-order relation knowledge from a pretrained high-resolution teacher model to a low-resolution student model. Through extensive experiments on low-resolution object classification and low-resolution face recognition, we validate the effectiveness and adaptability of our approach. Our future work will concentrate on integrating domain generalization and exploring its applicability to a broader spectrum of visual understanding tasks.

\myPara{Acknowledgements.}~This work was partially supported by grants from the National Key Research and Development Plan (2020AAA0140001) and Beijing Natural Science Foundation (19L2040).

\bibliographystyle{IEEEtran}
\bibliography{bibieee}

\begin{thebibliography}{10}
\providecommand{\url}[1]{#1}
\csname url@samestyle\endcsname
\providecommand{\newblock}{\relax}
\providecommand{\bibinfo}[2]{#2}
\providecommand{\BIBentrySTDinterwordspacing}{\spaceskip=0pt\relax}
\providecommand{\BIBentryALTinterwordstretchfactor}{4}
\providecommand{\BIBentryALTinterwordspacing}{\spaceskip=\fontdimen2\font plus
\BIBentryALTinterwordstretchfactor\fontdimen3\font minus
  \fontdimen4\font\relax}
\providecommand{\BIBforeignlanguage}[2]{{%
\expandafter\ifx\csname l@#1\endcsname\relax
\typeout{** WARNING: IEEEtran.bst: No hyphenation pattern has been}%
\typeout{** loaded for the language `#1'. Using the pattern for}%
\typeout{** the default language instead.}%
\else
\language=\csname l@#1\endcsname
\fi
#2}}
\providecommand{\BIBdecl}{\relax}
\BIBdecl

\bibitem{jia2021scaling}
C.~Jia, Y.~Yang, Y.~Xia, Y.-T. Chen, Z.~Parekh, H.~Pham, Q.~Le, Y.-H. Sung,
  Z.~Li, and T.~Duerig, ``Scaling up visual and vision-language representation
  learning with noisy text supervision,'' in \emph{Proceedings of the
  International Conference on Machine Learning (ICML)}, 2021, pp. 4904--4916.

\bibitem{huang2022tcsvt}
T.~Huang, X.~Ben, C.~Gong, B.~Zhang, R.~Yan, and Q.~Wu, ``Enhanced
  spatial-temporal salience for cross-view gait recognition,'' \emph{IEEE
  Transactions on Circuits and Systems for Video Technology (TCSVT)}, vol.~32,
  no.~10, pp. 6967--6980, 2022.

\bibitem{imagenet_cvpr09}
J.~Deng, W.~Dong, R.~Socher, L.-J. Li, K.~Li, and L.~Fei-Fei, ``Imagenet: A
  large-scale hierarchical image database,'' in \emph{IEEE Conference on
  Computer Vision and Pattern Recognition (CVPR)}, 2009, pp. 248--255.

\bibitem{ben2022tackling}
X.~Ben, C.~Gong, T.~Huang, C.~Li, R.~Yan, and Y.~Li, ``Tackling
  micro-expression data shortage via dataset alignment and active learning,''
  \emph{IEEE Transactions on Multimedia (TMM)}, pp. 1--14, 2022.

\bibitem{Groupface}
Y.~Kim, W.~Park, M.-C. Roh, and J.~Shin, ``Groupface: Learning latent groups
  and constructing group-based representations for face recognition,'' in
  \emph{IEEE/CVF Conference on Computer Vision and Pattern Recognition (CVPR)},
  2020, pp. 5620--5629.

\bibitem{LFWTech}
G.~B. Huang, M.~Ramesh, T.~Berg, and E.~Learned-Miller, ``Labeled faces in the
  wild: A database for studying face recognition in unconstrained
  environments,'' University of Massachusetts, Amherst, Tech. Rep., 2007.

\bibitem{ge2020tcsvt}
S.~Ge, C.~Li, S.~Zhao, and D.~Zeng, ``Occluded face recognition in the wild by
  identity-diversity inpainting,'' \emph{IEEE Transactions on Circuits and
  Systems for Video Technology (TCSVT)}, vol.~30, no.~10, pp. 3387--3397, 2020.

\bibitem{yan2022tcsvt}
T.~Yan, H.~Li, B.~Sun, Z.~Wang, and Z.~Luo, ``Discriminative feature mining and
  enhancement network for low-resolution fine-grained image recognition,''
  \emph{IEEE Transactions on Circuits and Systems for Video Technology
  (TCSVT)}, vol.~32, no.~8, pp. 5319--5330, 2022.

\bibitem{ge2018low}
S.~{Ge}, S.~{Zhao}, C.~{Li}, and J.~{Li}, ``Low-resolution face recognition in
  the wild via selective knowledge distillation,'' \emph{IEEE Transactions on
  Image Processing (TIP)}, vol.~28, no.~4, pp. 2051--2062, 2019.

\bibitem{GeHORKD20}
S.~Ge, K.~Zhang, H.~Liu, Y.~Hua, S.~Zhao, X.~Jin, and H.~Wen, ``Look one and
  more: Distilling hybrid order relational knowledge for cross-resolution image
  recognition,'' in \emph{Proceedings of the AAAI Conference on Artificial
  Intelligence (AAAI)}, 2020, pp. 10\,845--10\,852.

\bibitem{ge2020efficient}
S.~Ge, S.~Zhao, C.~Li, Y.~Zhang, and J.~Li, ``Efficient low-resolution face
  recognition via bridge distillation,'' \emph{IEEE Transactions on Image
  Processing (TIP)}, vol.~29, pp. 6898--6908, 2020.

\bibitem{beyer2022cvpr}
L.~Beyer, X.~Zhai, A.~Royer, L.~Markeeva, R.~Anil, and A.~Kolesnikov,
  ``Knowledge distillation: A good teacher is patient and consistent,'' in
  \emph{IEEE/CVF Conference on Computer Vision and Pattern Recognition (CVPR)},
  2022, pp. 10\,915--10\,924.

\bibitem{wang2016studying}
Z.~Wang, S.~Chang, Y.~Yang, D.~Liu, and T.~S. Huang, ``Studying very low
  resolution recognition using deep networks,'' in \emph{IEEE Conference on
  Computer Vision and Pattern Recognition (CVPR)}, 2016, pp. 4792--4800.

\bibitem{peng2019correlation}
B.~Peng, X.~Jin, J.~Liu, D.~Li, Y.~Wu, Y.~Liu, S.~Zhou, and Z.~Zhang,
  ``Correlation congruence for knowledge distillation,'' in \emph{IEEE/CVF
  International Conference on Computer Vision (ICCV)}, 2019, pp. 5007--5016.

\bibitem{tung2019similarity}
F.~Tung and G.~Mori, ``Similarity-preserving knowledge distillation,'' in
  \emph{IEEE/CVF International Conference on Computer Vision (ICCV)}, 2019, pp.
  1365--1374.

\bibitem{park2019relational}
W.~Park, D.~Kim, Y.~Lu, and M.~Cho, ``Relational knowledge distillation,'' in
  \emph{IEEE/CVF Conference on Computer Vision and Pattern Recognition (CVPR)},
  2019, pp. 3967--3976.

\bibitem{dong2021few}
S.~Dong, X.~Hong, X.~Tao, X.~Chang, X.~Wei, and Y.~Gong, ``Few-shot
  class-incremental learning via relation knowledge distillation,'' in
  \emph{Proceedings of the AAAI Conference on Artificial Intelligence (AAAI)},
  2021, pp. 1255--1263.

\bibitem{oord2018representation}
\BIBentryALTinterwordspacing
A.~Van~den Oord, Y.~Li, and O.~Vinyals, ``Representation learning with
  contrastive predictive coding,'' \emph{arXiv preprint}, 2018. [Online].
  Available: \url{https://arxiv.org/abs/1807.03748}
\BIBentrySTDinterwordspacing

\bibitem{tian2019crd}
\BIBentryALTinterwordspacing
Y.~Tian, D.~Krishnan, and P.~Isola, ``Contrastive representation
  distillation,'' in \emph{International Conference on Learning Representations
  (ICLR)}, 2020. [Online]. Available:
  \url{https://openreview.net/pdf?id=SkgpBJrtvS}
\BIBentrySTDinterwordspacing

\bibitem{xu2020knowledge}
G.~Xu, Z.~Liu, X.~Li, and C.~C. Loy, ``Knowledge distillation meets
  self-supervision,'' in \emph{European Conference on Computer Vision (ECCV)},
  2020, pp. 588--604.

\bibitem{zhu2021complementary}
J.~Zhu, S.~Tang, D.~Chen, S.~Yu, Y.~Liu, M.~Rong, A.~Yang, and X.~Wang,
  ``Complementary relation contrastive distillation,'' in \emph{IEEE/CVF
  Conference on Computer Vision and Pattern Recognition (CVPR)}, 2021, pp.
  9260--9269.

\bibitem{zou2011very}
W.~W. Zou and P.~C. Yuen, ``Very low resolution face recognition problem,''
  \emph{IEEE Transactions on image processing (TIP)}, vol.~21, no.~1, pp.
  327--340, 2011.

\bibitem{li2019low}
P.~Li, L.~Prieto, D.~Mery, and P.~J. Flynn, ``On low-resolution face
  recognition in the wild: Comparisons and new techniques,'' \emph{IEEE
  Transactions on Information Forensics and Security (TIFS)}, vol.~14, no.~8,
  pp. 2000--2012, 2019.

\bibitem{munir2021resolution}
A.~Munir, C.~Lyu, B.~Goossens, W.~Philips, and C.~Micheloni, ``Resolution based
  feature distillation for cross resolution person re-identification,'' in
  \emph{IEEE/CVF International Conference on Computer Vision (ICCV)}, 2021, pp.
  281--289.

\bibitem{mahapatra2019image}
D.~Mahapatra, B.~Bozorgtabar, and R.~Garnavi, ``Image super-resolution using
  progressive generative adversarial networks for medical image analysis,''
  \emph{Computerized Medical Imaging and Graphics}, vol.~71, pp. 30--39, 2019.

\bibitem{rouhsedaghat2021low}
M.~Rouhsedaghat, Y.~Wang, S.~Hu, S.~You, and C.-C.~J. Kuo, ``Low-resolution
  face recognition in resource-constrained environments,'' \emph{Pattern
  Recognition Letters}, vol. 149, pp. 193--199, 2021.

\bibitem{grm2019face}
K.~Grm, W.~J. Scheirer, and V.~{\v{S}}truc, ``Face hallucination using cascaded
  super-resolution and identity priors,'' \emph{IEEE Transactions on Image
  Processing (TIP)}, vol.~29, pp. 2150--2165, 2020.

\bibitem{9578478}
K.~C. Chan, X.~Wang, X.~Xu, J.~Gu, and C.~C. Loy, ``Glean: Generative latent
  bank for large-factor image super-resolution,'' in \emph{IEEE/CVF Conference
  on Computer Vision and Pattern Recognition (CVPR)}, 2021, pp.
  14\,240--14\,249.

\bibitem{kong2019cross}
\BIBentryALTinterwordspacing
H.~Kong, J.~Zhao, X.~Tu, J.~Xing, S.~Shen, and J.~Feng, ``Cross-resolution face
  recognition via prior-aided face hallucination and residual knowledge
  distillation,'' \emph{arXiv preprint}, 2019. [Online]. Available:
  \url{https://arxiv.org/abs/1905.10777}
\BIBentrySTDinterwordspacing

\bibitem{biswas2011multidimensional}
S.~Biswas, K.~W. Bowyer, and P.~J. Flynn, ``Multidimensional scaling for
  matching low-resolution face images,'' \emph{IEEE Transactions on Pattern
  Analysis and Machine Intelligence (TPAMI)}, vol.~34, no.~10, pp. 2019--2030,
  2011.

\bibitem{zangeneh2020low}
E.~Zangeneh, M.~Rahmati, and Y.~Mohsenzadeh, ``Low resolution face recognition
  using a two-branch deep convolutional neural network architecture,''
  \emph{Expert Systems with Applications}, vol. 139, p. 112854, 2020.

\bibitem{zha2019tcn}
J.~Zha and H.~Chao, ``Tcn: Transferable coupled network for cross-resolution
  face recognition,'' in \emph{Proceedings of the International Conference on
  Acoustics, Speech and Signal Processing (ICASSP)}, 2019, pp. 3302--3306.

\bibitem{Hinton2015DistillingTK}
\BIBentryALTinterwordspacing
G.~Hinton, O.~Vinyals, and J.~Dean, ``Distilling the knowledge in a neural
  network,'' in \emph{Advances in Neural Information Processing Systems
  Workshop on Deep Learning and Representation Learning}, 2015. [Online].
  Available: \url{http://arxiv.org/abs/1503.02531}
\BIBentrySTDinterwordspacing

\bibitem{gou2021knowledge}
J.~Gou, B.~Yu, S.~J. Maybank, and D.~Tao, ``Knowledge distillation: A survey,''
  \emph{International Journal of Computer Vision (IJCV)}, vol. 129, no.~6, pp.
  1789--1819, 2021.

\bibitem{liu2022tcsvt}
T.~Liu, K.~Lam, R.~Zhao, and G.~Qiu, ``Deep cross-modal representation learning
  and distillation for illumination-invariant pedestrian detection,''
  \emph{IEEE Transactions on Circuits and Systems for Video Technology
  (TCSVT)}, vol.~32, no.~1, pp. 315--329, 2022.

\bibitem{kim2017transferring}
\BIBentryALTinterwordspacing
S.~W. Kim and H.-E. Kim, ``Transferring knowledge to smaller network with
  class-distance loss,'' in \emph{International Conference on Learning
  Representations Workshop (ICLRW)}, 2017. [Online]. Available:
  \url{https://openreview.net/forum?id=ByXrfaGFe}
\BIBentrySTDinterwordspacing

\bibitem{mirzadeh2020improved}
S.~I. Mirzadeh, M.~Farajtabar, A.~Li, N.~Levine, A.~Matsukawa, and
  H.~Ghasemzadeh, ``Improved knowledge distillation via teacher assistant,'' in
  \emph{Proceedings of the AAAI Conference on Artificial Intelligence (AAAI)},
  vol.~34, no.~04, 2020, pp. 5191--5198.

\bibitem{zhao2022dkd}
B.~Zhao, Q.~Cui, R.~Song, Y.~Qiu, and J.~Liang, ``Decoupled knowledge
  distllation,'' in \emph{IEEE/CVF Conference on Computer Vision and Pattern
  Recognition (CVPR)}, 2022, pp. 11\,953--11\,962.

\bibitem{RomeroBKCGB14}
\BIBentryALTinterwordspacing
A.~Romero, N.~Ballas, S.~E. Kahou, A.~Chassang, C.~Gatta, and Y.~Bengio,
  ``Fitnets: Hints for thin deep nets,'' in \emph{International Conference on
  Learning Representations (ICLR)}, 2015. [Online]. Available:
  \url{https://arxiv.org/abs/1412.6550}
\BIBentrySTDinterwordspacing

\bibitem{heo2019comprehensive}
B.~Heo, J.~Kim, S.~Yun, H.~Park, N.~Kwak, and J.~Y. Choi, ``A comprehensive
  overhaul of feature distillation,'' in \emph{IEEE/CVF International
  Conference on Computer Vision (ICCV)}, 2019, pp. 1921--1930.

\bibitem{chen2021cross}
D.~Chen, J.-P. Mei, Y.~Zhang, C.~Wang, Z.~Wang, Y.~Feng, and C.~Chen,
  ``Cross-layer distillation with semantic calibration,'' in \emph{Proceedings
  of the AAAI Conference on Artificial Intelligence (AAAI)}, 2021, pp.
  7028--7036.

\bibitem{huang2022feature}
Z.~Huang, S.~Yang, M.~Zhou, Z.~Li, Z.~Gong, and Y.~Chen, ``Feature map
  distillation of thin nets for low-resolution object recognition,'' \emph{IEEE
  Transactions on Image Processing (TIP)}, vol.~31, pp. 1364--1379, 2022.

\bibitem{chen2021cvpr}
L.~Chen, D.~Wang, Z.~Gan, J.~Liu, R.~Henao, and L.~Carin, ``Wasserstein
  contrastive representation distillation,'' in \emph{IEEE/CVF Conference on
  Computer Vision and Pattern Recognition (CVPR)}, 2021, pp. 16\,296--16\,305.

\bibitem{he2020momentum}
K.~He, H.~Fan, Y.~Wu, S.~Xie, and R.~Girshick, ``Momentum contrast for
  unsupervised visual representation learning,'' in \emph{IEEE/CVF Conference
  on Computer Vision and Pattern Recognition (CVPR)}, 2020, pp. 9729--9738.

\bibitem{chen2020simple}
T.~Chen, S.~Kornblith, M.~Norouzi, and G.~Hinton, ``A simple framework for
  contrastive learning of visual representations,'' in \emph{Proceedings of the
  International Conference on Machine Learning (ICML)}, 2020, pp. 1597--1607.

\bibitem{hjelm2018learning}
\BIBentryALTinterwordspacing
R.~D. Hjelm, A.~Fedorov, S.~Lavoie-Marchildon, K.~Grewal, P.~Bachman,
  A.~Trischler, and Y.~Bengio, ``Learning deep representations by mutual
  information estimation and maximization,'' \emph{arXiv preprint}, 2018.
  [Online]. Available: \url{https://arxiv.org/abs/1808.06670}
\BIBentrySTDinterwordspacing

\bibitem{wu2018unsupervised}
Z.~Wu, Y.~Xiong, S.~X. Yu, and D.~Lin, ``Unsupervised feature learning via
  non-parametric instance discrimination,'' in \emph{IEEE/CVF Conference on
  Computer Vision and Pattern Recognition (CVPR)}, 2018, pp. 3733--3742.

\bibitem{zheng2021boosting}
K.~Zheng, Y.~Wang, and Y.~Yuan, ``Boosting contrastive learning with relation
  knowledge distillation,'' in \emph{Proceedings of the AAAI Conference on
  Artificial Intelligence (AAAI)}, 2022, pp. 3508--3516.

\bibitem{krizhevsky2009learning}
A.~Krizhevsky, ``Learning multiple layers of features from tiny images,''
  University of Toronto, Tech. Rep., 2009.

\bibitem{netzer2011reading}
Y.~Netzer, T.~Wang, A.~Coates, A.~Bissacco, B.~Wu, and A.~Y. Ng, ``Reading
  digits in natural images with unsupervised feature learning,'' in
  \emph{Advances in Neural Information Processing Systems Workshop}, 2011, pp.
  1--9.

\bibitem{Coates2011aistats}
A.~Coates, A.~Y. Ng, and H.~Lee, ``An analysis of single-layer networks in
  unsupervised feature learning,'' in \emph{International Conference on
  Artificial Intelligence and Statistics (AISTATS)}, 2011, pp. 215--223.

\bibitem{TinyImageNet}
L.~Hansen, ``Tiny imagenet challenge submission,'' in \emph{CS 231N}, 2015.

\bibitem{yi2014learning}
\BIBentryALTinterwordspacing
D.~Yi, Z.~Lei, S.~Liao, and S.~Z. Li, ``Learning face representation from
  scratch,'' \emph{arXiv preprint}, 2014. [Online]. Available:
  \url{https://arxiv.org/abs/1411.7923}
\BIBentrySTDinterwordspacing

\bibitem{uccs}
M.~{Günther}, P.~{Hu}, C.~{Herrmann} \emph{et~al.}, ``Unconstrained face
  detection and open-set face recognition challenge,'' in \emph{International
  Joint Conference on Biometrics (IJCB)}, 2017, pp. 697--706.

\bibitem{cheng2018accv}
Z.~Cheng, X.~Zhu, and S.~Gong, ``Low-resolution face recognition,'' in
  \emph{Asian Conference on Computer Vision (ACCV)}, 2018, pp. 605--621.

\bibitem{SimonyanZ14a}
\BIBentryALTinterwordspacing
K.~Simonyan and A.~Zisserman, ``Very deep convolutional networks for
  large-scale image recognition,'' in \emph{International Conference on
  Learning Representations (ICLR)}, 2015. [Online]. Available:
  \url{http://arxiv.org/abs/1409.1556}
\BIBentrySTDinterwordspacing

\bibitem{HeZRS16}
K.~He, X.~Zhang, S.~Ren, and J.~Sun, ``Deep residual learning for image
  recognition,'' in \emph{IEEE Conference on Computer Vision and Pattern
  Recognition (CVPR)}, 2016, pp. 770--778.

\bibitem{ZagoruykoK16}
S.~Zagoruyko and N.~Komodakis, ``Wide residual networks,'' in \emph{Proceedings
  of the British Machine Vision Conference (BMVC)}, 2016, pp. 87.1--87.12.

\bibitem{zhang2018shufflenet}
X.~Zhang, X.~Zhou, M.~Lin, and J.~Sun, ``Shufflenet: An extremely efficient
  convolutional neural network for mobile devices,'' in \emph{IEEE/CVF
  Conference on Computer Vision and Pattern Recognition (CVPR)}, 2018, pp.
  6848--6856.

\bibitem{ma2018shufflenet}
N.~Ma, X.~Zhang, H.-T. Zheng, and J.~Sun, ``Shufflenet v2: Practical guidelines
  for efficient cnn architecture design,'' in \emph{European Conference on
  Computer Vision (ECCV)}, 2018, pp. 116--131.

\bibitem{ZagoruykoK17}
\BIBentryALTinterwordspacing
N.~K. Zagoruyko, ``Paying more attention to attention: Improving the
  performance of convolutional neural networks via attention transfer,'' in
  \emph{International Conference on Learning Representations (ICLR)}, 2017.
  [Online]. Available: \url{https://openreview.net/forum?id=Sks9_ajex}
\BIBentrySTDinterwordspacing

\bibitem{pkt_eccv}
N.~Passalis and A.~Tefas, ``Learning deep representations with probabilistic
  knowledge transfer,'' in \emph{European Conference on Computer Vision
  (ECCV)}, 2018, pp. 268--284.

\bibitem{ahn2019variational}
S.~Ahn, S.~X. Hu, A.~Damianou, N.~D. Lawrence, and Z.~Dai, ``Variational
  information distillation for knowledge transfer,'' in \emph{IEEE/CVF
  Conference on Computer Vision and Pattern Recognition (CVPR)}, 2019, pp.
  9163--9171.

\bibitem{heo2019knowledge}
B.~Heo, M.~Lee, S.~Yun, and J.~Y. Choi, ``Knowledge transfer via distillation
  of activation boundaries formed by hidden neurons,'' in \emph{Proceedings of
  the AAAI Conference on Artificial Intelligence (AAAI)}, 2019, pp. 3779--3787.

\bibitem{KimPK18}
K.~Jangho, P.~Seonguk, and K.~Nojun, ``Paraphrasing complex network: Network
  compression via factor transfer,'' in \emph{Advances in Neural Information
  Processing Systems}, 2018, pp. 2765--2774.

\bibitem{zhang2021student}
K.~Zhang, C.~Zhanga, S.~Li, D.~Zeng, and S.~Ge, ``Student network learning via
  evolutionary knowledge distillation,'' \emph{IEEE Transactions on Circuits
  and Systems for Video Technology (TCSVT)}, vol.~32, no.~4, pp. 2251--2263,
  2022.

\bibitem{singh2021derivenet}
M.~Singh, S.~Nagpal, R.~Singh, and M.~Vatsa, ``Derivenet for (very) low
  resolution image classification,'' \emph{IEEE Transactions on Pattern
  Analysis and Machine Intelligence (TPAMI)}, vol.~44, no.~10, pp. 6569--6577,
  2022.

\bibitem{zhang2018super}
K.~Zhang, Z.~Zhang, C.-W. Cheng, W.~H. Hsu, Y.~Qiao, W.~Liu, and T.~Zhang,
  ``Super-identity convolutional neural network for face hallucination,'' in
  \emph{European Conference on Computer Vision (ECCV)}, 2018, pp. 183--198.

\bibitem{singh2019dual}
M.~Singh, S.~Nagpal, R.~Singh, and M.~Vatsa, ``Dual directed capsule network
  for very low resolution image recognition,'' in \emph{IEEE/CVF International
  Conference on Computer Vision (ICCV)}, 2019, pp. 340--349.

\bibitem{parkhi2015deep}
O.~Parkhi, A.~Vedaldi, and A.~Zisserman, ``Deep face recognition,''
  \emph{British Machine Vision Conference (BMVC)}, pp. 41.1--41.12, 2015.

\bibitem{sun2015deep}
Y.~Sun, Y.~Chen, X.~Wang, and X.~Tang, ``Deep learning face representation by
  joint identification-verification,'' in \emph{Advances in Neural Information
  Processing Systems}, 2014, pp. 1988--1996.

\bibitem{schroff2015facenet}
F.~Schroff, D.~Kalenichenko, and J.~Philbin, ``Facenet: A unified embedding for
  face recognition and clustering,'' in \emph{IEEE Conference on Computer
  Vision and Pattern Recognition (CVPR)}, 2015, pp. 815--823.

\bibitem{luo2016face}
P.~Luo, Z.~Zhu, Z.~Liu, X.~Wang, and X.~Tang, ``Face model compression by
  distilling knowledge from neurons,'' in \emph{Proceedings of the AAAI
  Conference on Artificial Intelligence (AAAI)}, 2016, pp. 3560--3566.

\bibitem{liu2017sphereface}
W.~Liu, Y.~Wen, Z.~Yu, M.~Li, B.~Raj, and L.~Song, ``Sphereface: Deep
  hypersphere embedding for face recognition,'' in \emph{IEEE Conference on
  Computer Vision and Pattern Recognition (CVPR)}, 2017, pp. 212--220.

\bibitem{wu2018shift}
B.~Wu, A.~Wan, X.~Yue, P.~Jin, S.~Zhao, N.~Golmant, A.~Gholaminejad,
  J.~Gonzalez, and K.~Keutzer, ``Shift: A zero flop, zero parameter alternative
  to spatial convolutions,'' in \emph{IEEE/CVF Conference on Computer Vision
  and Pattern Recognition (CVPR)}, 2018, pp. 9127--9135.

\bibitem{wang2018cosface}
H.~Wang, Y.~Wang, Z.~Zhou, X.~Ji, D.~Gong, J.~Zhou, Z.~Li, and W.~Liu,
  ``Cosface: Large margin cosine loss for deep face recognition,'' in
  \emph{IEEE/CVF Conference on Computer Vision and Pattern Recognition (CVPR)},
  2018, pp. 5265--5274.

\bibitem{Cao2018VGGFace2}
Q.~Cao, L.~Shen, W.~Xie, O.~M. Parkhi, and A.~Zisserman, ``Vggface2: A dataset
  for recognising faces across pose and age,'' in \emph{IEEE International
  Conference on Automatic Face Gesture Recognition (FG)}, 2018, pp. 67--74.

\bibitem{deng2019arcface}
J.~Deng, J.~Guo, N.~Xue, and S.~Zafeiriou, ``Arcface: Additive angular margin
  loss for deep face recognition,'' in \emph{IEEE/CVF Conference on Computer
  Vision and Pattern Recognition (CVPR)}, 2019, pp. 4690--4699.

\bibitem{meng2021magface}
Q.~Meng, S.~Zhao, Z.~Huang, and F.~Zhou, ``Magface: A universal representation
  for face recognition and quality assessment,'' in \emph{IEEE/CVF Conference
  on Computer Vision and Pattern Recognition (CVPR)}, 2021, pp.
  14\,225--14\,234.

\bibitem{li2022deep}
P.~Li, S.~Tu, and L.~Xu, ``Deep rival penalized competitive learning for
  low-resolution face recognition,'' \emph{Neural Networks}, vol. 148, pp.
  183--193, 2022.

\bibitem{talreja2019attribute}
V.~Talreja, F.~Taherkhani, M.~C. Valenti, and N.~M. Nasrabadi,
  ``Attribute-guided coupled gan for cross-resolution face recognition,'' in
  \emph{IEEE International Conference on Biometrics Theory, Applications and
  Systems (BTAS)}, 2019, pp. 1--10.

\bibitem{wen2016discriminative}
Y.~Wen, K.~Zhang, Z.~Li, and Y.~Qiao, ``A discriminative feature learning
  approach for deep face recognition,'' in \emph{European Conference on
  Computer Vision (ECCV)}, 2016, pp. 499--515.

\bibitem{lu2018deep}
Z.~Lu, X.~Jiang, and A.~Kot, ``Deep coupled resnet for low-resolution face
  recognition,'' \emph{IEEE Signal Processing Letters}, pp. 526--530, 2018.

\bibitem{li2018face}
\BIBentryALTinterwordspacing
P.~Li, L.~Prieto, D.~Mery, and P.~Flynn, ``Face recognition in low quality
  images: a survey,'' \emph{arXiv preprint}, 2018. [Online]. Available:
  \url{https://arxiv.org/abs/1805.11519}
\BIBentrySTDinterwordspacing

\bibitem{van2008visualizing}
L.~Van~der Maaten and G.~Hinton, ``Visualizing data using t-sne,''
  \emph{Journal of Machine Learning Research (JMLR)}, vol.~9, no.~86, pp.
  2579--2605, 2008.

\end{thebibliography}

\begin{IEEEbiography}[{\includegraphics[width=1in,height=1.25in,clip,keepaspectratio]{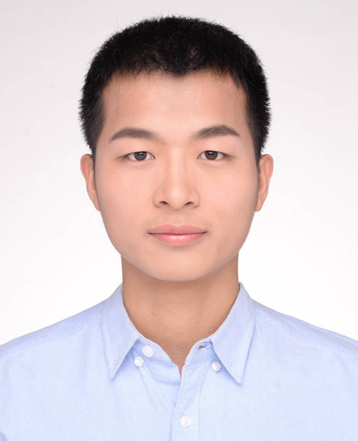}}]{Kangkai Zhang} received his B.S. degree in Electronical Information Science and Technology from the School of Electronic Information and Optical Engineering in Nankai University, Tianjin, China. He obtained a Master's degree in Communication and Information Systems at the Institute of Information Engineering at Chinese Academy of Sciences, Beijing. Currently, he works as a Computer Vision Algorithm Engineer at Baidu Inc. His major research interests are deep learning and computer vision.
\end{IEEEbiography}

\begin{IEEEbiography}[{\includegraphics[width=1in,height=1.25in,clip,keepaspectratio]{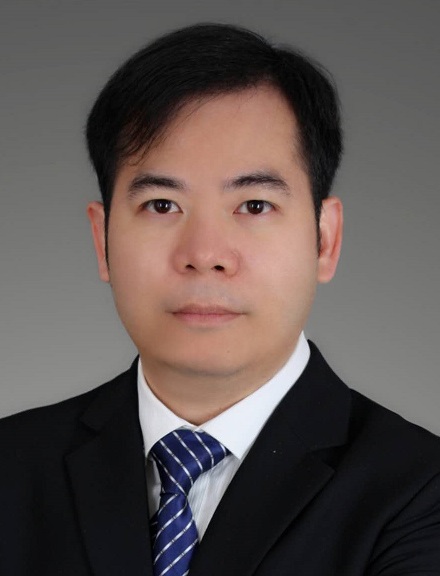}}]{Shiming Ge} (M'13-SM'15) is a professor with the Institute of Information Engineering, Chinese Academy of Sciences. Prior to that, he was a senior researcher and project manager in Shanda Innovations, a researcher in Samsung Electronics and Nokia Research Center. He received the B.S. and Ph.D degrees both in Electronic Engineering from the University of Science and Technology of China (USTC) in 2003 and 2008, respectively. His research mainly focuses on computer vision, data analysis, machine learning and AI security, especially efficient and trustworthy solutions towards scalable applications. He is a senior member of IEEE, CSIG and CCF.
\end{IEEEbiography}

\begin{IEEEbiography}[{\includegraphics[width=1in,height=1.25in,clip,keepaspectratio]{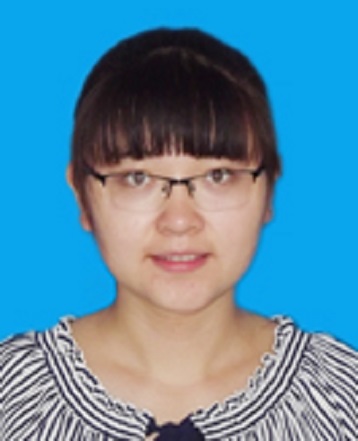}}]{Ruixin Shi} received her B.S. degree in Information Security from the School of Cyber Security in Beijing Institute of Technology, China. She is now a Ph.D Candidate at the Institute of Information Engineering at Chinese Academy of Sciences and the School of Cyber Security at the University of Chinese Academy of Sciences, Beijing. His major research interests are computer vision and generative modeling.
\end{IEEEbiography}

\begin{IEEEbiography}[{\includegraphics[width=1in,height=1.25in,clip,keepaspectratio]{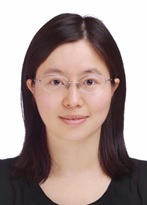}}]{Dan Zeng} (SM'21) received her Ph.D. degree in circuits and systems, and her B.S. degree in electronic science and technology, both from University of Science and Technology of China, Hefei. She is a full professor and the Dean of the Department of Communication Engineering at Shanghai University, directing the Computer Vision and Pattern Recognition Lab. Her main research interests include computer vision, multimedia analysis, and machine learning. She is serving as the Associate Editor of the IEEE Transactions on Multimedia and the IEEE Transactions on Circuits and Systems for Video Technology, the TC Member of IEEE MSA and Associate TC member of IEEE MMSP.
\end{IEEEbiography}

\vfill

\end{document}